\documentclass[runningheads]{llncs}
\usepackage{graphicx}
\usepackage{tikz}
\usepackage{comment}
\usepackage{amsmath,amssymb} % define this before the line numbering.
\usepackage{color}

\usepackage[accsupp]{axessibility}  % Improves PDF readability for those with disabilities.

\usepackage{booktabs}

\usepackage{epsfig}
\usepackage{xspace}
\usepackage{blindtext}
\usepackage{caption}
\usepackage{amsfonts}
\usepackage[inline]{enumitem}
\usepackage{algorithmic}
\usepackage{textcomp}

\usepackage{multirow}
\usepackage{adjustbox}
\usepackage{etoolbox}
\usepackage{dsfont}

\usepackage{float}
\usepackage{lipsum}
\usepackage{stfloats}
\usepackage{multicol}
\usepackage{bm}
\usepackage{array}
\usepackage{tabulary}
\usepackage{paralist}

\usepackage{bbding}
\usepackage{pifont}
\newcommand{\cmark}{\ding{51}}%
\newcommand{\xmark}{\ding{55}}%
\usepackage{colortbl}
\usepackage{diagbox}
\usepackage{wrapfig}
\usepackage{caption}
\usepackage[normalem]{ulem}
\usepackage{subfig}

\usepackage[pagebackref,breaklinks,colorlinks]{hyperref}

\newcommand{\vB}{\mathbf{B}}

\newcommand{\vL}{\mathbf{L}}
\definecolor{tableHeadGray}{gray}{.9}

\makeatletter
\DeclareRobustCommand\onedot{\futurelet\@let@token\@onedot}
\def\@onedot{\ifx\@let@token.\else.\null\fi\xspace}
 
\def\ie{\emph{i.e}\onedot} 
 
\def\etc{\emph{etc}\onedot} 
 
\def\etal{\emph{et al}\onedot}
\makeatother

\usepackage[capitalize]{cleveref}
\crefname{section}{Sec.}{Secs.}
\Crefname{section}{Section}{Sections}
\Crefname{table}{Table}{Tables}
\crefname{table}{Tab.}{Tabs.}

\newcommand{\PAR}[1]{\noindent{\bf #1}}

\DeclareMathOperator*{\signum}{sgn}
\DeclareMathOperator*{\softmax}{softmax}
\DeclareMathOperator*{\attention}{Attention}

\newcommand{\name}{0}
\newcommand{\h}{0}
\newcommand{\w}{0.15}
\newcommand{\wa}{0.15}
\newlength \g

\begin{document}
% \linenumbers
\pagestyle{headings}
\mainmatter
\def\ECCVSubNumber{}  
\title{Event-Based Fusion for Motion Deblurring with Cross-modal Attention}

% \begin{comment}
\titlerunning{Event-Based Fusion for Motion Deblurring with Cross-modal Attention}
\author{Lei Sun\inst{1,2} \and
Christos Sakaridis\inst{2} \and
Jingyun Liang\inst{2} \and
Qi Jiang\inst{1} \and
Kailun Yang\inst{3} \and
Peng Sun\inst{1} \and
Yaozu Ye\inst{1} \and
Kaiwei Wang\inst{1} \and
Luc Van Gool\inst{2,4}
}
\authorrunning{L. Sun et al.}
\institute{Zhejiang University \and
ETH Zürich \and
Karlsruhe Institute of Technology \and
KU Leuven}
% \end{comment}
%******************
\maketitle


%%%%%%%%% ABSTRACT
\begin{abstract}
   Traditional frame-based cameras inevitably suffer from motion blur due to long exposure times. As a kind of bio-inspired camera, the event camera records the intensity changes in an asynchronous way with high temporal resolution, providing valid image degradation information within the exposure time. In this paper, we rethink the event-based image deblurring problem and unfold it into an end-to-end two-stage image restoration network. To effectively fuse event and image features, we design an event-image cross-modal attention module applied at multiple levels of our network, which allows to focus on relevant features from the event branch and filter out noise. We also introduce a novel symmetric cumulative event representation specifically for image deblurring as well as an event mask gated connection between the two stages of our network which helps avoid information loss. At the dataset level, to foster event-based motion deblurring and to facilitate evaluation on challenging real-world images, we introduce the Real Event Blur (REBlur) dataset, captured with an event camera in an illumination-controlled optical laboratory. Our Event Fusion Network (EFNet) sets the new state of the art in motion deblurring, surpassing both the prior best-performing image-based method and all event-based methods with public implementations on the GoPro dataset (by up to 2.47dB) and on our REBlur dataset, even in extreme blurry conditions. The code and our REBlur dataset will be made publicly available.
\end{abstract}

%%%%%%%%% BODY TEXT
\section{Introduction}
\label{sec:intro}

Motion blur often occurs in images due to camera shake or object motion during the exposure time. The goal of deblurring is to recover a sharp image with clear edge structures and texture details from the blurry image. This is a highly ill-posed problem because of the infinitely many feasible solutions~\cite{bahat2017non,cho2009fast,zhang2017beyond}. Traditional methods explicitly utilize natural image priors and various constraints~\cite{bahat2017non,fergus2006removing,kotera2013blind,krishnan2011blind,levin2009understanding,levin2011efficient,xu2013unnatural}. To better generalize when addressing the deblurring problem, modern learning-based methods choose to train Convolutional Neural Networks (CNNs) on large-scale data to learn the implicit relationships between blurry and sharp images~\cite{gong2017motion,nah2017deep,sun2015learning,tao2018scale,zhang2018dynamic}. Despite their high performance on existing public datasets, these learning-based methods often fail when facing extreme or real-world blur. Their performance heavily relies on the quality and scale of the training data, which creates the need for a more general and reliable deblurring method.

Event cameras~\cite{brandli2014240,gallego2020event,patrick2008128x} are bio-inspired asynchronous sensors with high temporal resolution (in the order of $\mu$s) and they operate well in environments with high dynamic range. Different from traditional frame-based cameras, event cameras capture the intensity change of each pixel (i.e.\ \emph{event} information) independently, if the change surpasses a threshold. Event cameras encode the intensity change information within the exposure time of the image frame into an event stream, making it possible to deblur an image frame with events~\cite{pan2019bringing_high_framerate}. However, because of sensor noise and uncertainty in the aforementioned threshold, it is difficult to use a physical model to deblur images based solely on events. Thus, some methods~\cite{jiang2020learning_event_motion_deblur,lin2020learning_event_video_deblur,shang2021bringing} utilize CNNs to deal with noise corruption and threshold uncertainty. Nevertheless, these methods only achieve slight performance gains compared to image-only methods, due to rather ineffective event representations and fusion mechanisms between events and images.

In this paper, we first revisit the mechanism of motion blur and how event information is utilized in image reconstruction. To deal with the inherent defect of the event-based motion deblurring equation, we propose EFNet, an Event Fusion Network for image deblurring which effectively combines information from event and frame-based cameras for image deblurring. Motivated by the physical model of event-based image deblurring~\cite{pan2019bringing_high_framerate}, we design a symmetric cumulative event representation (SCER) specifically for deblurring and formulate our network based on a two-stage image restoration model. Each stage of the model has a U-Net-like architecture~\cite{ronneberger2015u}. The first stage consists of two branches, an image branch and an event branch, the features of which are fused at multiple levels. In order to perform the fusion of the two modalities, we propose an Event-Image Cross-modal Attention (EICA) fusion module, which allows to attend to the event features that are relevant for deblurring via a channel-level attention mechanism. To the best of our knowledge, this is the first time that a multi-head attention mechanism is applied to event-based image deblurring. We also enable information exchange between the two stages of our network by applying Event Mask Gated Connections (EMGC), which selectively transfer feature maps from the encoder and decoder of the first stage to the second stage. A detailed ablation study shows the effectiveness of our novel fusion module using cross-modal attention, our gated connection module and our multi-level middle fusion design. Additionally, we record a real-world event blur dataset named Real Event Blur (REBlur) in an optical laboratory with stable illumination and a high-precision electronically controlled slide-rail which allows various types of motion. We conduct extensive comparisons against state-of-the-art deblurring methods on the GoPro dataset~\cite{nah2017deep} with synthetic events and on REBlur with real events and demonstrate the superiority of our event-based image deblurring method.
% By fine-tuning on the training set of the REBlur dataset, we bridge the gap between simulated events and real-world events, and help our model generalize to the real world better.

In summary, we make the following main contributions:
\begin{compactitem}
    \item We design a novel event-image fusion module which applies cross-modal channel-wise attention to adaptively fuse event features with image features, and incorporate it at multiple levels of a novel end-to-end deblurring network.
    \item We introduce a novel symmetric cumulative event voxel representation for deblurring, which is inspired by the physical model that connects blurry image formation and event generation.
    \item We present REBlur, a real-world dataset consisting of tuples of blurry images, sharp images and event streams from an event camera, which provides a challenging evaluation setting for deblurring methods.
    \item Our deblurring network, equipped with our proposed modules and event representation, sets the new state of the art for image deblurring on the GoPro dataset and our REBlur dataset.
\end{compactitem}

\section{Related Work}
\label{sec:related_work}

\PAR{Image deblurring.} Traditional approaches often formulate deblurring as an optimization problem~\cite{fergus2006removing,kotera2013blind,krishnan2011blind,levin2009understanding,levin2011efficient,xu2013unnatural}. Recently, with the success of deep learning, 
image deblurring has achieved impressive performance thanks to the usage of CNNs. CNN-based methods directly map the blurry image to the latent sharp image. Several novel components and techniques have been proposed, such as attention modules~\cite{suin2020spatially,tsai2021banet}, multi-scale fusion~\cite{nah2017deep,tao2018scale}, multi-stage networks~\cite{chen2021hinet,zamir2021multi}, and coarse-to-fine strategies~\cite{cho2021rethinking}, improving the accuracy and robustness of deblurring. Despite the benefits they have shown for deblurring, all aforementioned deep networks operate solely on images, a modality which does not explicitly capture \emph{motion} and thus inherently limits performance when facing real-world blurry images, especially in extreme conditions.

\PAR{Event-based deblurring.} Recently, events have been used for motion deblurring, due to the strong connection they possess with motion information. Pan~\etal~\cite{pan2019bringing_high_framerate} proposed an Event Double Integral (EDI) deblurring model using the double integral of event data. They established a mathematical event-based model mapping blurry frames to sharp frames, which is a seminal approach to deblurring with events. However, limited by the sampling mechanism of event cameras, this method often introduces strong, accumulated noise.
Jiang~\etal~\cite{jiang2020learning_event_motion_deblur} extracted motion information and sharp edges from events to assist deblurring. However, their early fusion approach, which merely concatenates events into the main branch of the network, does not take full advantage of events. Lin~\etal~\cite{lin2020learning_event_video_deblur} fused events with the image via dynamic filters from STFAN~\cite{zhou2019spatio}. In addition, Shang~\etal~\cite{shang2021bringing} fused event information into a weight matrix that can be applied to any state-of-the-art network.
To sum up, most of the above event-based learning methods did not use event information effectively, achieving only minor improvements compared to image-only methods on standard benchmarks.

\PAR{Event representation.} Different from synchronous signals such as images from frame-based cameras, events are asynchronous and sparse. A key point in how to extract information from events effectively is the representation of the events. Event representation is an application-dependent problem and different tasks admit different solutions. The event-by-event method is suitable for spiking neural networks owing to its asynchronous architecture~\cite{snnref2,snnref3,snnref1}. A Time Surface, which is a 2D map that stores the time value deriving from the timestamp of the last event, has proved suitable for event-based classification~\cite{ahad2012motion,lagorce2016hots,sironi2018hats}.
Some modern learning-based methods convert events to a 2D frame by counting events or accumulating polarities~\cite{liu2018adaptive,maqueda2018event,shang2021bringing}. This approach is compatible with conventional computer vision tasks but loses temporal information. 3D space-time histograms of events, also called voxel grids, preserve the temporal information of events better by accumulating event polarity on a voxel~\cite{bardow2016simultaneous,zhu2019evflownet}. For image deblurring, most works utilized 2D event-image pairs~\cite{shang2021bringing} or borrowed 3D voxel grids like Stacking Based on Time (SBT) from image reconstruction~\cite{wang2019event}. However, there still is no event representation specifically designed for motion deblurring.

\section{Method}
\label{sec:method}

\begin{figure*}[tb]
    % \captionsetup{font=small}
    \centering 
    \includegraphics[width=1.0\textwidth]{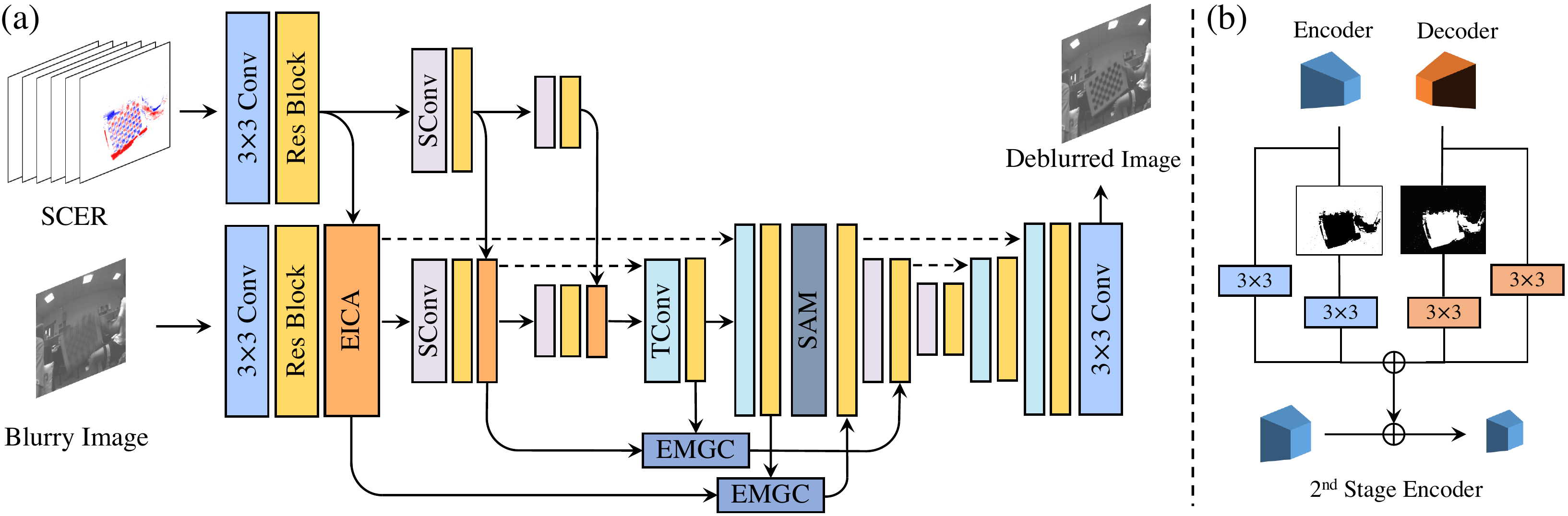}
    \vspace{-5mm}
    \caption{(a): \textbf{The architecture of our Event Fusion Network (EFNet).} EFNet consists of two UNet-like backbones~\cite{ronneberger2015u} and an event extraction branch. After each residual convolution block (``Res Block''), feature maps from the event branch and the image branch are fused. The second UNet backbone refines the deblurred image further. ``SCER": symmetric cumulative event representation, ``EICA'': event-image cross-modal attention, ``SConv": 4$\times$4 strided convolution with stride 2, ``TConv": 2$\times$2 transposed convolution with stride 2, ``SAM'': supervision attention module~\cite{zamir2021multi}. (b): \textbf{The Event Mask Gated Connection module (EMGC)} transfers features across stages guided by an event mask.}
    \vspace{-5mm}
    \label{fig:model}
\end{figure*}

We first introduce the mathematical model for the formation of blurry images from sharp images that involves events in Sec.~\ref{sec:method:theory}. Based on this model, we pose the event-based deblurring problem as a deblurring-denoising problem and base the high-level design of our network architecture on this formulation, as explained in Sec.~\ref{sec:method:general}. We present our symmetric cumulative representation for events, which constitutes a 3D voxel grid in which the temporal dimension is discretized, in Sec.~\ref{sec:method:event:representation}. This event representation is provided as input together with the blurry image to our two-stage network. We then detail the two main novel components of our network: our novel event-image cross-modal attention fusion mechanism (Sec.~\ref{sec:method:eica}), which adaptively fuses feature channels associated with events and images, and our event mask gated connection module between the two stages of our network (Sec.~\ref{sec:method:emgc}), which helps selectively forward to the second stage the features at sharp regions of the input from the encoder and the features at blurry regions from the decoder of the first stage.

\subsection{Problem Formulation}
\label{sec:method:theory}
% if this part contains too many, divide it to a new section

% SHORT INTRODUCE TO EVENT TRIGGER CONDITIONS.
For an event camera, the $i$-th event $e_i$ is represented as a tuple $e_i = (x_i, y_i, t_i, p_i)$, where $x_i$, $y_i$ and $t_i$ represent the pixel coordinates and the timestamp of the event respectively, and $p_i\in \left \{ -1, +1  \right \}$ is the polarity of the event~\cite{brandli2014240,patrick2008128x}. An event is triggered at time $t$ only when the change in pixel intensity $\mathcal{I}$ surpasses the threshold compared to the pixel intensity at the time of the last trigger. This is formulated as 
\begin{equation}
\label{eq:event_polarity}
p_i = 
\begin{cases}
+1, \text{if} \log \left( \frac{\mathcal{I}_t {(x_i, y_i)}} {\mathcal{I}_{ t - \Delta t} {(x_i, y_i)} } \right) > c, \\
-1, \text{if} \log \left( \frac{\mathcal{I}_t {(x_i, y_i)}} {\mathcal{I}_{ t - \Delta t} {(x_i, y_i)} } \right) < -c,
\end{cases}
\end{equation}
where $c$ is the contrast threshold of intensity change, which may differ across the sensor plane.

Given the intensity of a latent sharp image $\vL$, according to~\cite{pan2019bringing_high_framerate}, the corresponding blurred image $\vB$ can be derived by the Event-based Double Integral (EDI) model:
{\small
\begin{equation} \label{eq:edi}
\begin{split}
\vB &=  \frac{1}{T} \int_{f-T/2}^{f+T/2} \vL(t) dt\\
% &= \frac{\vL(f)}{T}\int_{f-T/2}^{f+T/2} \widehat{c}^{\vE(t)} dt\\
% & = \vL(f) \ \frac{1}{T}\int_{f-T/2}^{f+T/2} \exp (c\ \vE(t) )\ dt\\
& = \frac{\vL(f)}{T}\int_{f-T/2}^{f+T/2}  \exp \Big (c \int_{f}^{t} p(s) ds\Big)\ dt,
\end{split}
\end{equation}
}where $f$ is the middle point of the exposure time $T$, $p(s)$ is the polarity component of the event stream and $\vL(f)$ is the latent sharp image corresponding to the blurred image $\vB$. The discretized version of \eqref{eq:edi} can be expressed as
{\small
\begin{equation} \label{eq:discre_edi}
\vB = \frac{\vL(N)}{2N+1} \sum_{i=0}^{2N} \exp\left(c \signum(i-N) \sum_{j:\;m \leq t_j \leq M} p_j\delta_{x_j y_j}\right),
\end{equation}
}where $\signum$ is the signum function, $m = \min\{f+T/2(i/N-1),f\}$, $M = \max\{f+T/2(i/N-1),f\}$ and $\delta$ is the Kronecker delta, defined as
\begin{equation} \label{eq:kronecker:delta}
    \delta_{kl}(m,n) = 
    \begin{cases}
    1, \text{ if } k=m \text{ and } l=n,\\
    0, \text{ otherwise.}
    \end{cases}
\end{equation}
In \eqref{eq:discre_edi}, we partition the exposure time $T$ into $2N$ equal intervals. Rearranging \eqref{eq:discre_edi} yields:
{\small
\begin{equation}\label{eq:discre_ln}
\vL(N) = \frac{(2N+1)\vB}{\sum_{i=0}^{2N} \exp\left(c \signum(i-N) \sum_{j:\;m \leq t_j \leq M} p_j\delta_{x_j y_j}\right)}.
\end{equation}
}

\subsection{General Architecture of EFNet}
\label{sec:method:general}

The formulation in \eqref{eq:discre_ln} indicates that the latent sharp image can be derived from the blurred image combined with the set of events $\mathcal{E} = \{e_i = (x_i, y_i, t_i, p_i): f - T/2 \leq t_i \leq f + T/2\}$ (\ie, all the events which are triggered within the exposure time), when events in this set are accumulated over time. We propose to learn this relation with a deep neural network, named Event Fusion Network (EFNet), which admits as inputs the blurred image and the events and maps them to the sharp image. The generic form of the learned mapping is
\begin{equation}
\label{eq:simplify_deblur_1}
    \vL_{\text{initial}} = f_3\left(f_{1}(\vB; \Theta_{1}), f_{2}(\mathcal{E};\Theta_{2}); \Theta_3\right),
\end{equation}
where the blurred image and the events are mapped individually to intermediate representations via $f_1$ and $f_2$ respectively and these intermediate representations are afterwards passed to a joint mapping $f_3$. $\Theta_1$, $\Theta_2$ and $\Theta_3$ denote the respective parameters of the three mappings. The main challenges we need to address given this generic formulation of our model are (i) how to represent the set of events $\mathcal{E}$ in a suitable way for inputting it to the network, and (ii) how and when to fuse the intermediate representations that are generated for the blurred image by $f_1$ and for the events by $f_2$, \ie, how to design $f_3$. We address the issue of how to represent the events in Sec.~\ref{sec:method:event:representation} and how to perform fusion in Sec.~\ref{sec:method:eica}.

\eqref{eq:discre_edi} is the ideal formulation for event-based motion image deblurring. However, in real-world settings, three factors make it impossible to restore the image simply based on this equation:
% (1): Instead of being strictly equal to a fixed value, the values of threshold $c$ for a given event camera are neither constant in time nor across the image space~\cite{stoffregen2020reducing,wang19acra}. (2): Intensity changes that are lower than the threshold $c$ do not trigger an event. (3): Spurious events occur over the entire image.

\begin{compactitem}
    \item{Instead of being strictly equal to a fixed value, the values of threshold $c$ for a given event camera are neither constant in time nor across the image~\cite{stoffregen2020reducing,wang19acra}}.
    \item{Intensity changes that are lower than the threshold $c$ do not trigger an event.}
    \item{Spurious events occur over the entire image.}
\end{compactitem}

Most of the restoration errors come from the first two factors, which result in degradation of the restored image in regions with events. We denote these regions as $R_e$.
%The rest of the image ${R^{\prime}_{e}}$ is much less affected by spurious events and minor intensity changes.
Taking the above factors into account, we design our network so that it includes a final mapping  of the initial deblurred image $\vL_{\text{initial}}$ to a denoised version of it, which can correct potential errors in the values of pixels inside $R_e$:
\begin{equation}
\label{eq:simplify_deblur_2}
    \vL_{\text{final}} = f_4(\vL_{\text{initial}};\Theta_4).
\end{equation}

% \begin{wrapfigure}[t]
%   \centering
%   \captionsetup{font=small}
%   \tiny
% %   \fbox{\rule{0pt}{2in} \rule{0.9\linewidth}{0pt}}
%   \includegraphics[width=0.5\linewidth]{./figures/SCER.pdf}
%   \vspace{-3.5mm}
%   \caption{\textbf{The proposed Symmetric Cumulative Event Representation (SCER).} Red and blue dots represent events with positive and negative polarity respectively. The bottom part of the figure depicts the individual channels of SCER and the saturation of the dot color represents the relative intensity change corresponding to the central latent sharp frame.}
%   \label{fig:event_representation}
%   \vspace{-4mm}
% \end{wrapfigure}

% Module names: Affine event-image fusion module, Event Mask Gated Connection module (EMGC).
% Network name: Event Fusion Network (EFNet)?
% Between two stages, we borrow a supervised attention module (SAM) from~\cite{zamir2021multi} to help rescale the feature maps before second stage.
\PAR{Two-stage backbone.} Based on the generic architecture of the network presented in \eqref{eq:simplify_deblur_1} and \eqref{eq:simplify_deblur_2}, we design EFNet as a two-stage network to progressively restore sharp images from blurred images and event streams, where the first and second stage implement the mappings in \eqref{eq:simplify_deblur_1} and \eqref{eq:simplify_deblur_2} respectively. The detailed architecture of EFNet is illustrated in Fig.~\ref{fig:model}. Both stages of EFNet have an encoder-decoder structure, based on the UNet~\cite{ronneberger2015u} architecture, and each stage consists of two down-sampling and two up-sampling layers. Between the encoder and decoder, we add a skip connection with $3\times3$ convolution. The residual convolution block in the UNet consists of two $3\times3$ convolution layers and leaky ReLU functions with a $1\times1$ convolution shortcut. Recently, the Supervised Attention Module (SAM) in multi-stage methods demonstrated superior capacity in transferring features between different sub-networks~\cite{chen2021hinet,zamir2021multi}. Thus, we use SAM to connect the two stages of our network. In the first stage, we fuse features from the event branch and the image branch at multiple levels using a novel cross-modal attention-based block. Between the two stages, we design an Event Mask Gated Connection module to boost feature aggregation with blurring priors from events. The details of the two aforementioned components of EFNet are given in Sec.~\ref{sec:method:eica} and \ref{sec:method:emgc}.

\subsection{Symmetric Cumulative Event Representation}
\label{sec:method:event:representation}

\begin{wrapfigure}{l}{0.5\textwidth}
  \vspace{-30pt}
  \begin{center}
   \includegraphics[width=0.48\textwidth]{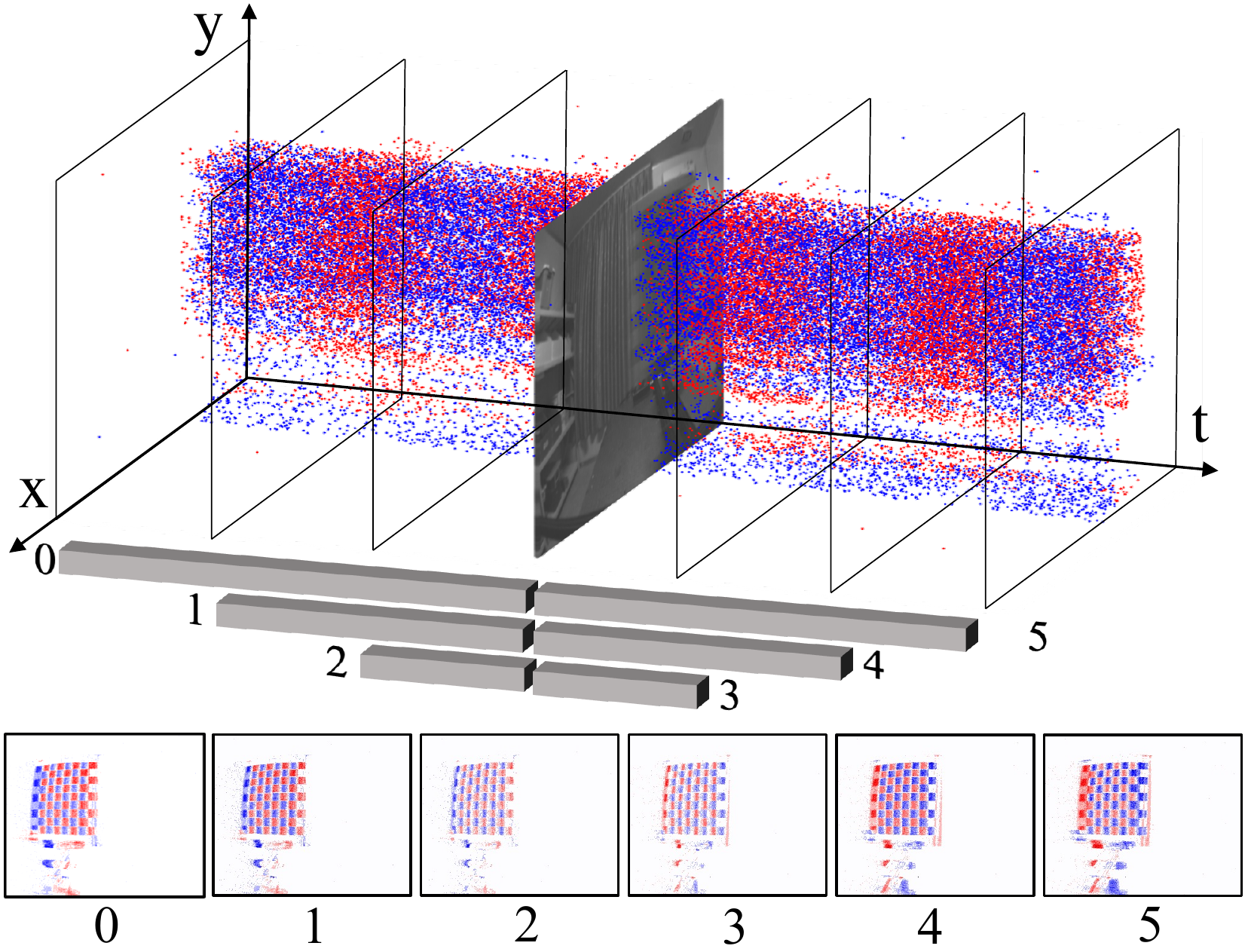}
  \end{center}
  \vspace{-10pt}
  % \captionsetup{font={small}}
  \caption{\textbf{The proposed Symmetric Cumulative Event Representation (SCER).} Red and blue dots represent events with positive and negative polarity respectively. }
%   The bottom part of the figure depicts the individual channels of SCER and the saturation of the dot color represents the relative intensity change corresponding to the central latent sharp frame.}
  \vspace{-17pt}
  \label{fig:event_representation}
\end{wrapfigure}

To feed the asynchronous events corresponding to synchronized image frames to our network, we design a representation specifically suited for deblurring. In \eqref{eq:discre_edi}, the accumulation of polarities via the inner sum on the right-hand side indicates the relative intensity changes between the target latent sharp image $\vL(N)$ and each of the rest of latent sharp images in the exposure time. The accumulation via the outer sum on the right-hand side represents the sum of all latent sharp images. Based on this relationship, we propose the Symmetric Cumulative Event Representation (SCER). As Fig.~\ref{fig:event_representation} shows, the exposure time $T$ of the blurry image is divided equally into $2N$ intervals. Assuming $2N+1$ latent sharp images in $T$, the polarity accumulation from the central target latent image $\vL(N)$ to a single latent image turns into a 2D tensor with dimensions $(H,W)$:
\begin{equation} \label{eq:SCER}
\mathbf{SCER}_i = \signum(i-N) \sum_{j:\;m \leq t_j \leq M} p_j\delta_{x_j y_j}.
\end{equation}

For $i=N$, $\mathbf{SCER}_N = 0$, so we discard this tensor. The remaining $2N$ tensors are concatenated together, forming a tensor which indicates intensity changes between the central latent sharp image $\vL(N)$ and each of the $2N$ other latent images. In this way, $\mathbf{SCER} \in \mathbb{R}^{H\times W \times 2N}$ includes all the relative intensity values corresponding to the center latent sharp frame and it becomes suitable for feature extraction with our image deblurring model. As the accumulation limits change, our SCER also contains both information about the area in which blur occurs (channel 0 and channel $2N-1$) and information about sharp edges (channel $N-1$ and channel $N$).

Our method discretizes $T$ into $2N$ parts, quantizing temporal information of events within the time interval $\frac{T}{2N}$. However, SCER still holds temporal information, as the endpoints of the time interval in which events are accumulated is different across channels. The larger $N$ is, the less temporal information is lost. In our implementation, we fix $N=3$.

% \begin{wrapfigure}[行数]{位置}[超出长度]{宽度}
% 图形命令
% \end{wrapfigure}

\begin{wrapfigure}{l}{0.5\textwidth}
  \vspace{-30pt}
  \begin{center}
   \includegraphics[width=0.48\textwidth]{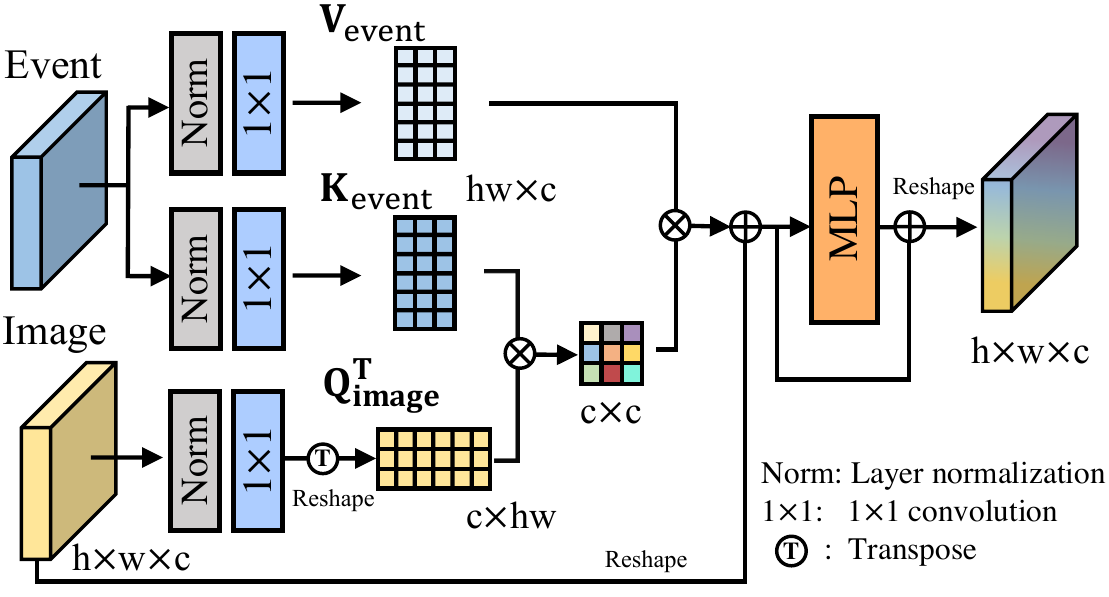}
  \end{center}
  \vspace{-10pt}
  \caption{\textbf{The proposed Event-Image Cross-modal Attention fusion module.} The size of the attention map is $c\times c$.}
  \vspace{-20pt}
  \label{fig:EICA}
\end{wrapfigure}

\subsection{Event-Image Cross-modal Attention Fusion}
\label{sec:method:eica}

How to jointly extract and fuse information from event streams and images is the key to event-based deblurring. Previous work~\cite{jiang2020learning_event_motion_deblur,lin2020learning_event_video_deblur} simply multiplies or concatenates low-resolution feature maps from the two modalities, but this simple fusion approach cannot fully model the relation between events and images. Other methods estimate optical flow with events and use that for deblurring~\cite{shang2021bringing}. However, the estimation of optical flow itself introduces errors.

To utilize event data, we instead include a novel cross-modal attention block at multiple levels of EFNet. Contrary to self-attention blocks, in which the queries ($\mathbf{Q}$), keys ($\mathbf{K}$) and values ($\mathbf{V}$) all come from the same branch of the network, our Event-Image Cross-modal Attention (EICA) block admits as inputs the queries $\mathbf{Q}_{\text{image}}$ from the image branch and the keys $\mathbf{K}_{\text{event}}$ and values $\mathbf{V}_{\text{event}}$ from the event branch, as shown in Fig.~\ref{fig:EICA}. In particular, the input features from the two branches are first fed to normalization and $1\times1$ convolution layers, where the latter have $c$ output channels. We then apply cross-modal attention between vectorized features from the two modalities via

\begin{equation} \label{eq:EICA}
  \attention(\mathbf{Q}_{\text{image}},\mathbf{K}_{\text{event}},\mathbf{V}_{\text{event}}) = \mathbf{V}_{\text{event}}\softmax\left(\frac{\mathbf{Q}^{T}_{\text{image}}\mathbf{K}_{\text{event}}}{\sqrt{d_{k}}}\right).
\end{equation}

The reason for introducing the $1\times1$ convolution layer is to reduce the spatial complexity of the above attention operation. In particular, $c$ is chosen to be much smaller than $hw$, where $h$ and $w$ are the height and width of the input feature maps, and the soft indexing of $\mathbf{K}_{\text{event}}$ by $\mathbf{Q}_{\text{image}}$ is performed at the \emph{channel} dimension instead of the spatial dimensions. Thus, the resulting soft attention map from \eqref{eq:EICA} is $c\times{}c$ instead of $hw\times{}hw$, reducing the spatial complexity from $\mathcal{O}(h^2w^2)$ to $\mathcal{O}(c^2)$ and making the operation feasible even for features with high spatial resolution, as in our case.
Finally, the output of the attention operation is added to the input image features and the result of this addition is passed to a  multi-layer perceptron (MLP) consisting of two fully connected layers with a Gaussian Error Linear Unit (GELU)~\cite{hendrycks2016gaussian} in between. We use the EICA module at multiple levels of EFNEt to fuse event information aggregated across receptive fields of varying size.

\subsection{Event Mask Gated Connection Module}
\label{sec:method:emgc}

Previous work~\cite{purohit2021spatially} predicts a mask indicating which areas of an image are severely distorted, but this mask is not completely accurate. Apart from information about intensity changes, event data also contain spatial information about the blurred regions of the input image. Typically, regions in which events occur are more severely degraded in the blurry image. Motivated by this observation, we introduce an Event Mask Gated Connection (EMGC) between the two stages of our network to exploit the spatial information about blurred regions.

In particular, we binarize the sum of the first and last channel of SCER and thus obtain a binary event mask, in which pixels where an event has occurred are set to 0 and the rest are set to 1. As illustrated in Fig.~\ref{fig:model}(b), EMGC masks out the feature maps of the encoder at regions where the event mask is 0, which are expected to be more blurry, and masks out the feature maps of the decoder at regions where the event mask is 1 (using the complement of the event mask), which are expected to be less blurry. A skip connection is added beside the mask operation. Feature maps with less artifacts in the encoder and better restored feature maps are combined through the event mask gate.
In this way, we selectively connect feature maps from the encoder and the decoder of the first stage to the second stage. Besides, EMGC eases the flow of information through the network, as it creates a shortcut through which features can be transferred directly from the first to the second stage.

\section{REBlur Dataset}
\label{sec:reblur_dataset}

Almost all event-based motion deblurring methods~\cite{haoyu2020learning,jiang2020learning_event_motion_deblur,lin2020learning_event_video_deblur,shang2021bringing,xu2021motion_deblur_events} train models on blurred image datasets, such as GoPro~\cite{nah2017deep}, with synthetic events from ESIM~\cite{rebecq2018esim}. 
Although the contrast threshold $c$ in the event simulator varies across pixels as in reality, a domain gap between synthetic and real events still exists because of the background activity noise, dark current noise, and false negatives in refractory period~\cite{baldwin2020event,stoffregen2020reducing,wang19acra}. Recently, Jiang~\etal~\cite{jiang2020learning_event_motion_deblur} proposed BlurDVS by capturing an image plus events with slow motion, and then synthesizing motion blur by averaging multiple nearby frames. However, motion blur in the ground-truth images is inevitable in this setting and fast motion causes different events from slow motion because of the false negatives in the refractory period of event cameras~\cite{baldwin2020event,xu2021motion_deblur_events}.
Thus, a large-scale real-world dataset with blurry images, reliable corresponding events, and ground-truth sharp images is missing.

% The temporal resolution of events is up to $10$kHz and we fix the exposure time of the electronic global shutter to $46$ms.
\sloppy{We present a new event-based dataset for deblurring, Real Event Blur (REBlur), to provide ground truth for blurry images in a two-shot way. To collect REBlur, we built an image collection system in a high-precision optical laboratory with very stable illumination. We fixed an Insightness Seem 1 event camera and a Dynamic and Active Pixel Vision Sensor (DAVIS) to the optical table, outputting time-aligned event streams and $260\times360$ gray images. To obtain blurry-sharp image pairs under high-speed motion, we also fixed a high-precision electronic-controlled slide-rail system to the optical table. In the first shot, we captured images with motion blur for the pattern on the slide-rail and corresponding event streams. In the second shot, according to the timestamp $t_{s}$ of the blurry images, we selected events within the time range $[t_{s} - 125\mu s, t_{s} + 125\mu s]$ and visualized these events in the preview of the sharp image capture program. Referring to the edge information from high-temporal-resolution events, we could relocate the slide-rail to the coordinate corresponding to the timestamp $t_{s}$ by an electronic-controlled stepping motor and then capture the latent sharp image. Between the two shots, the background was kept static.}

\begin{wrapfigure}{l}{5.5cm}
    \vspace{-30pt}
    \begin{center}
    \includegraphics[width=0.45\textwidth]{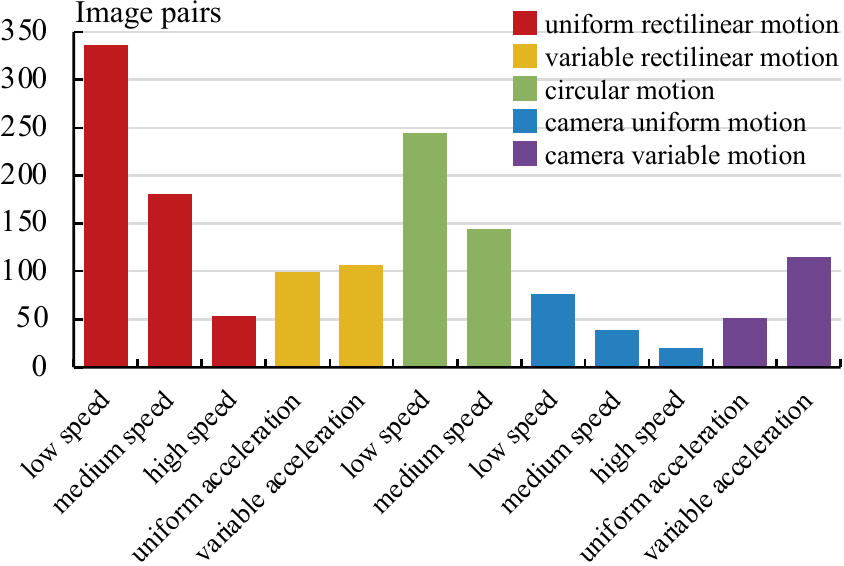}
    \end{center}
    \vspace{-10pt}
    \caption{\textbf{Distribution of different motion categories in our REBlur dataset.}}
    \vspace{-15pt}
    \label{fig:dataset}
\end{wrapfigure}

To enhance the generalization of the network for different objects and moving processes, our dataset includes 12 kinds of linear and nonlinear motions for $3$ different moving patterns and for the camera itself, as detailed in Fig.~\ref{fig:dataset}. The dataset consists of $36$ sequences and $1469$ groups of blurry-sharp image pairs with associated events, where $486$ pairs are used for training and $983$ for testing. We also include an additional set of 4 sequences including extreme blur, without ground truth. Please refer to the supplement for more details on REBlur.

\begin{table}[tb]
% \captionsetup{font=small}
% \vspace{-3mm}
\caption{\small \textbf{Comparison of motion deblurring methods on the GoPro dataset~\cite{nah2017deep}.} ${\dagger}$ denotes event-based methods. SRN+ and HINet+ denote event-enhanced versions of SRN~\cite{tao2018scale} and HINet~\cite{chen2021hinet} using our SCER.}
% \vspace{-3mm}
\centering
\small
% \tiny
\resizebox{0.95\textwidth}{!}{
\setlength{\tabcolsep}{30pt}     % adjust the width of cell
\renewcommand\arraystretch{1.0} % adjust hight of the row
\begin{tabular}{ l  ||  c  |  c  }
\bottomrule[0.15em]
\rowcolor{tableHeadGray}
 Method & PSNR $\uparrow$ & SSIM $\uparrow$  \\ \hline \hline
DeblurGAN \cite{kupyn2018deblurgan}                           & 28.70 {(54.1\%)} & 0.858 {(80.3\%)}   \\
BHA$^{\dagger}$~\cite{pan2019bringing_high_framerate}         & 29.06 {(52.1\%)} & 0.940 {(53.3\%)}  \\
Nah \etal \cite{nah2017deep}                                  & 29.08 {(52.0\%)} & 0.914 {(67.4\%)}   \\
DeblurGAN-v2 \cite{Kupyn_2019_ICCV}                           & 29.55 {(49.4\%)} & 0.934 {(57.6\%)}     \\
SRN~\cite{tao2018scale}                                       & 30.26 {(45.1\%)} & 0.934 {(57.6\%)}    \\
SRN+$^{\dagger}$~\cite{tao2018scale}                          & 31.02 {(40.0\%)} & 0.936 {(56.3\%)}   \\
DMPHN \cite{zhang2019deep}                                    & 31.20 {(38.8\%)} & 0.940 {(53.3\%)}   \\
D$^{2}$Nets$^{\dagger}$~\cite{shang2021bringing}              & 31.60 {(35.9\%)} & 0.940 {(53.3\%)} \\
LEMD$^{\dagger}$~\cite{jiang2020learning_event_motion_deblur} & 31.79 {(34.5\%)} & 0.949 {(45.1\%)}  \\
Suin \etal \cite{suin2020spatially}                           & 31.85 {(34.0\%)} & 0.948 {(46.2\%)}   \\
SPAIR\cite{purohit2021spatially}                              & 32.06 {(32.4\%)} & 0.953 {(40.4\%)}  \\ 
MPRNet~\cite{zamir2021multi}                                  & 32.66 {(27.6\%)} & 0.959 {(31.7\%)}   \\
HINet~\cite{chen2021hinet}                                    & 32.71 {(27.1\%)} & 0.959 {(31.7\%)}  \\
ERDNet$^{\dagger}$~\cite{haoyu2020learning}                   & 32.99 {(24.8\%)} & 0.935 {(56.9\%)}  \\
HINet+$^{\dagger}$~\cite{chen2021hinet}                       & 33.69 {(18.4\%)} & 0.961 {(28.2\%)}  \\

% \bottomrule[0.1em]
\hline
\textbf{EFNet (Ours)$^{\dagger}$}                           & \textbf{35.46} & \textbf{0.972}  \\
% \bottomrule
\hline
\end{tabular}
}
\label{table:gopro}
\vspace{-0.5cm}
\end{table}

\section{Experiments}

\subsection{Datasets and Settings}

\PAR{GoPro dataset.}
We use the GoPro dataset~\cite{nah2017deep}, which is widely used in motion deblurring, for training and evaluation. It consists of $3214$ pairs of blurry and sharp images with a resolution of $1280\times720$ and the blurred images are produced by averaging several high-speed sharp images. We use 2103 pairs for training and 1111 pairs for testing, following standard practice~\cite{nah2017deep}. We use ESIM~\cite{rebecq2018esim}, an open-source event camera simulator, to generate simulated event data for GoPro. To make the results more realistic, we set the contrast threshold $c$ randomly for each pixel, following a Gaussian distribution $\mathit{N}(\mu=0.2, \sigma=0.03)$.

\PAR{REBlur dataset.}
In order to close the gap between simulated events and real events, before evaluating models that are trained on GoPro on REBlur, we fine-tune them on the training set of REBlur. We then evaluate the fine-tuned models on the test set of REBlur. More details on this fine-tuning follow.

\PAR{Implementation details.}
Our network requires no pre-training. We train it on $256\times256$ crops of full images from GoPro. Full details about our network configuration (numbers of channels, kernel sizes \etc) are given in the supplement. For data augmentation, horizontal and vertical flipping, random noise and hot pixels in event voxels~\cite{stoffregen2020reducing} are applied. We use Adam~\cite{kingma2014adam} with an initial learning rate of $2\times10^{-4}$, and the cosine learning rate strategy with a minimum learning rate of $10^{-7}$. The model is trained with a batch size of $8$ for 300k iterations. Fine-tuning on REBlur involves $600$ iterations, the initial learning rate is $2\times10^{-5}$ and other configurations are kept the same as for GoPro. We use the same training and fine-tuning settings for our method and other methods for a fair comparison.

\PAR{Evaluation protocol.}
All quantitative comparisons are performed using PSNR and SSIM~\cite{wang2004image}. Apart from these, we also report the relative reduction in error of the best-performing model for the GoPro benchmark. This is done by converting PSNR to RMSE ($\textrm{RMSE} \propto \sqrt{10^{-\textrm{PSNR}/10}}$) and translating SSIM to DSSIM ($\textrm{DSSIM} = (1 - \textrm{SSIM})/2$).

\begin{figure*}[tb]
	% \captionsetup{font=small}
	\centering
% 	\scriptsize
    % \Large
	%\tiny
	%\scriptsize
	\footnotesize
% 	\small
	
	%%%%%
	\renewcommand{\h}{0.105}
	\renewcommand{\wa}{0.12}
	\newcommand{\wb}{0.16}
	\renewcommand{\g}{-0.7mm}
	\renewcommand{\tabcolsep}{1.8pt}
	\renewcommand{\arraystretch}{1}
	\resizebox{1.00\linewidth}{!} {
		%   	\hspace{-2mm}
		\begin{tabular}{cc}
			
			\renewcommand{\name}{figures/qualitative_gopro/0003_}
			\renewcommand{\h}{0.12}
			\renewcommand{\w}{0.2}
			\begin{tabular}{cc}
				%\tiny
				%\scriptsize
				%\footnotesize
				%\small
				% \hspace{-2.3mm}
				\begin{adjustbox}{valign=t}
					%\tiny
					\begin{tabular}{c}%left bottom right top
		         	\includegraphics[trim={0 0 416 0 },clip, width=0.354\textwidth]{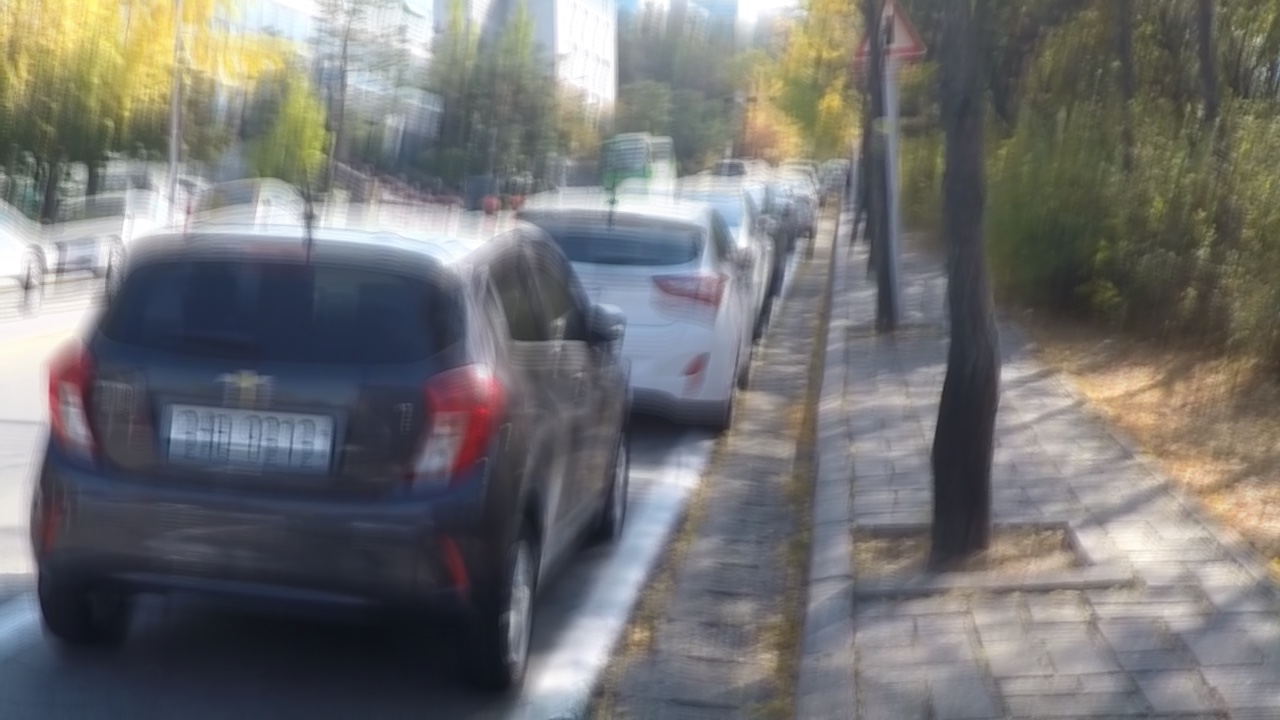}
						\\
						Blurry Image
					\end{tabular}
				\end{adjustbox}
				%		\hspace{-4.2mm}
				\begin{adjustbox}{valign=t}
					%\tiny
					\begin{tabular}{cccccc}
						\includegraphics[trim={158 230 935 378},clip,height=\h \textwidth, width=\w \textwidth]{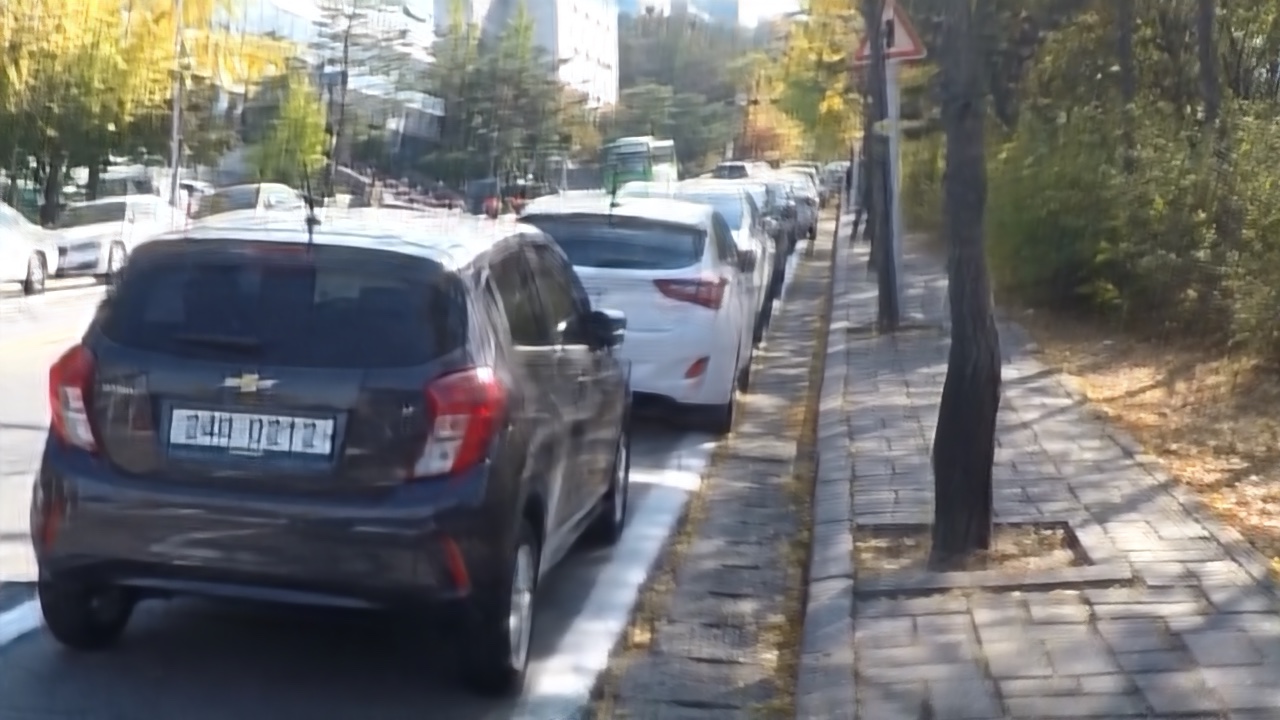} \hspace{\g} &
						\includegraphics[trim={158 230 935 378},clip,height=\h \textwidth, width=\w \textwidth]{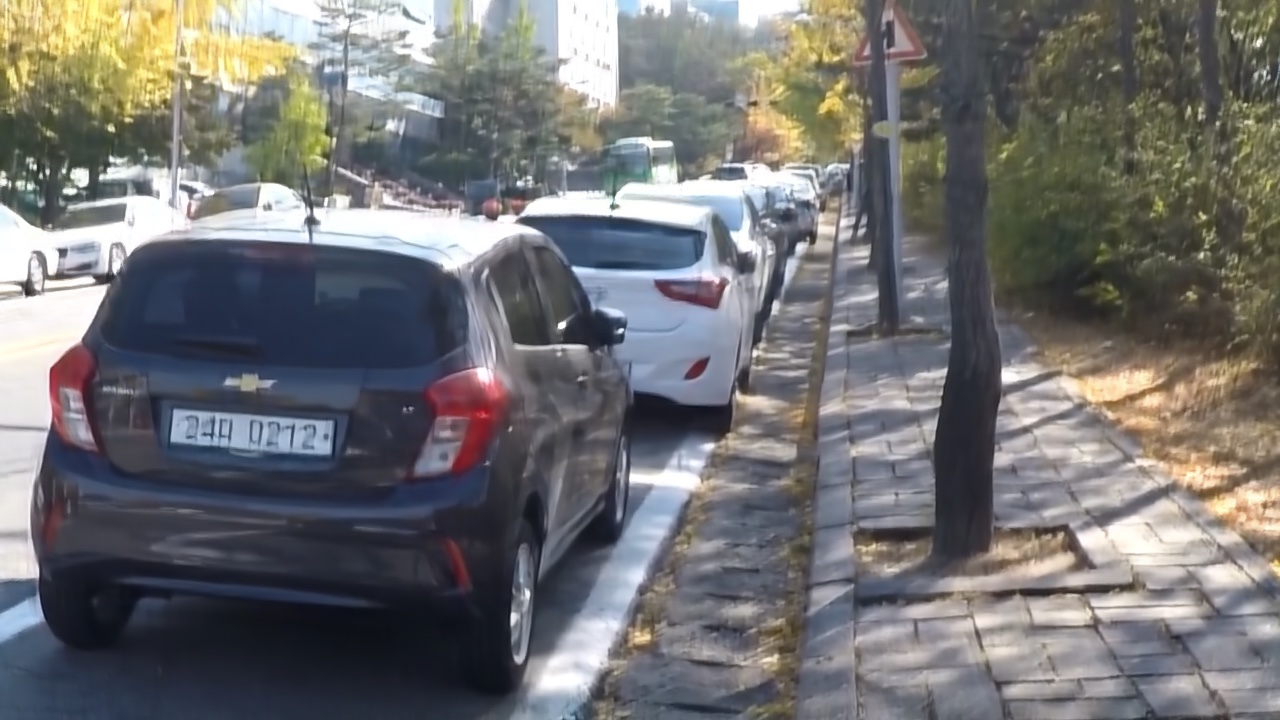} \hspace{\g} &
						\includegraphics[trim={158 230 935 378},clip,height=\h \textwidth, width=\w \textwidth]{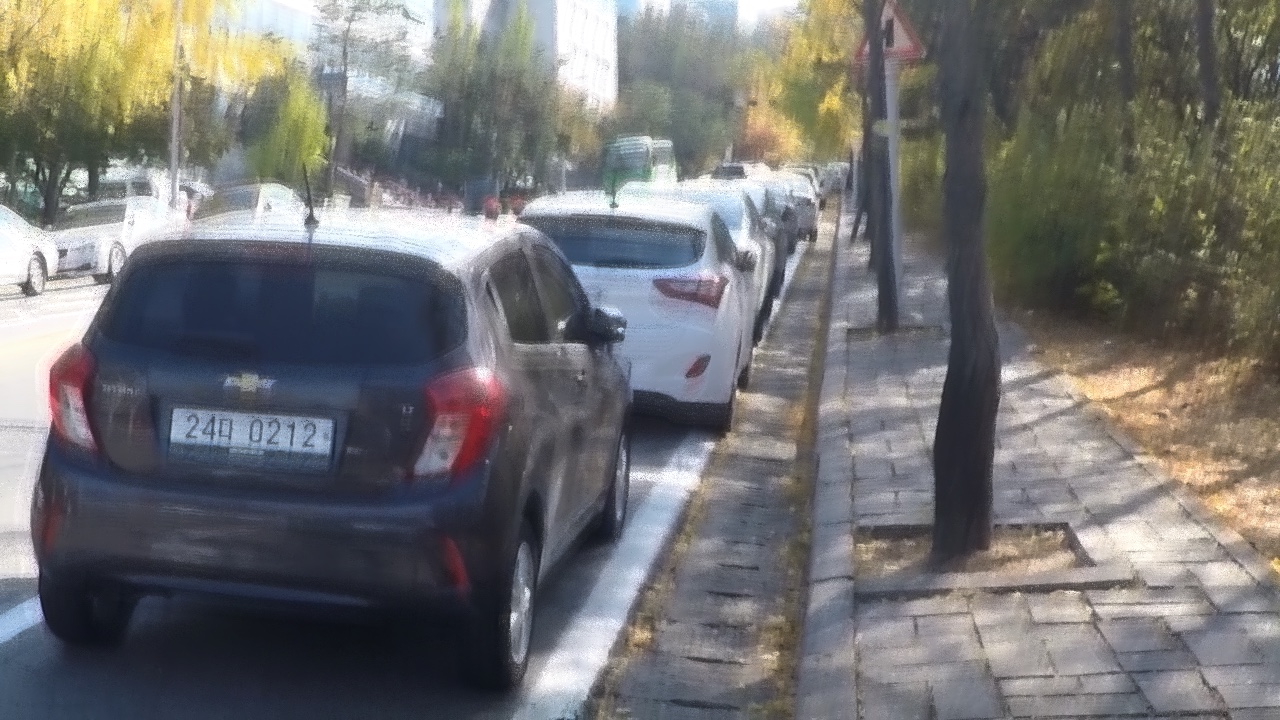} \hspace{\g} &
						\includegraphics[trim={158 230 935 378},clip,height=\h \textwidth, width=\w \textwidth]{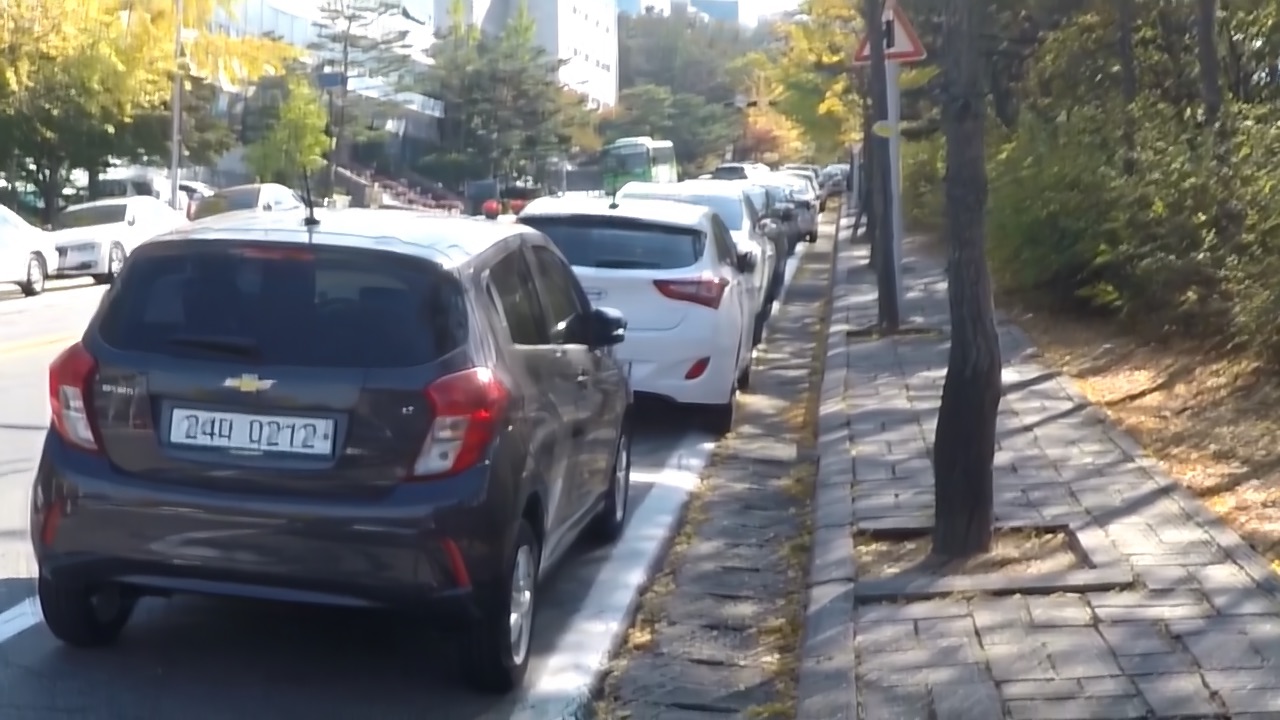} %MPRNet
						\\
						
						SRN~\cite{tao2018scale} &  HINet~\cite{chen2021hinet}  & BHA~\cite{pan2019bringing_high_framerate} &  MPRNet~\cite{zamir2021multi}
						\\
				% 		(a)
						\vspace{-2.45mm}
				% 		\vspace{-4mm}
						\\
						
						\includegraphics[trim={158 230 935 378},clip,height=\h \textwidth, width=\w \textwidth]{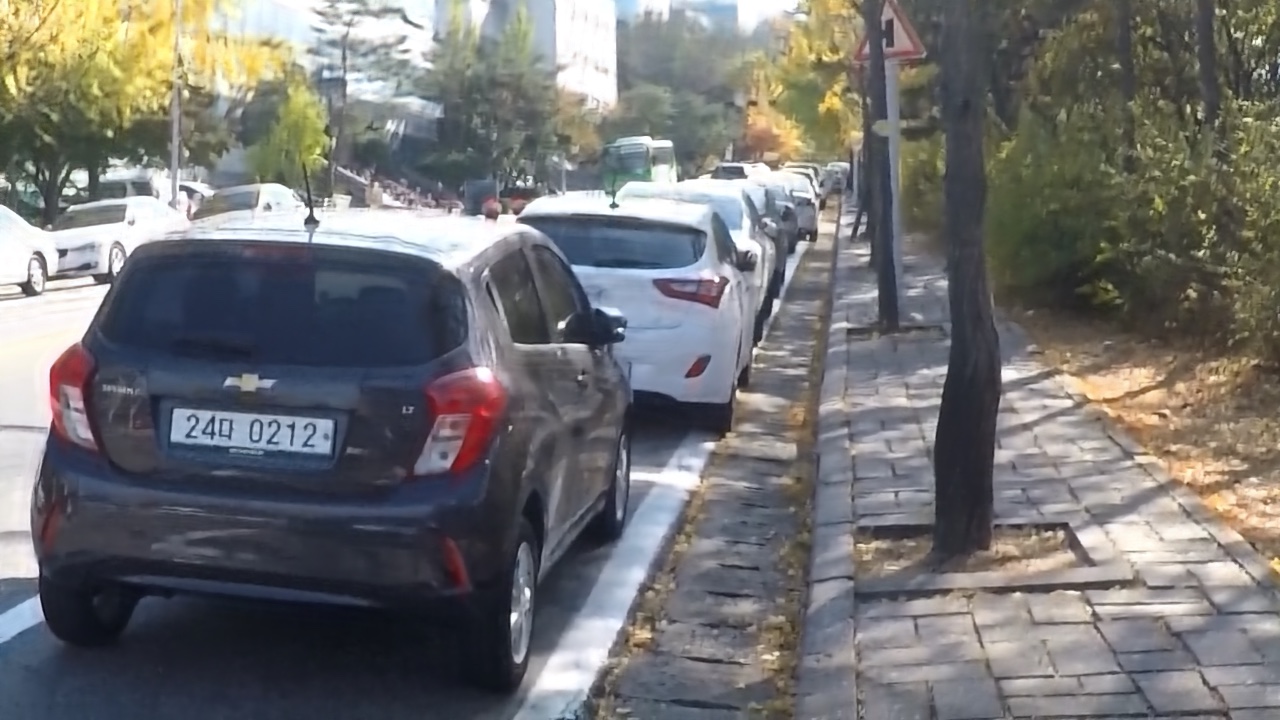} \hspace{\g} &
						\includegraphics[trim={158 230 935 378},clip,height=\h \textwidth, width=\w \textwidth]{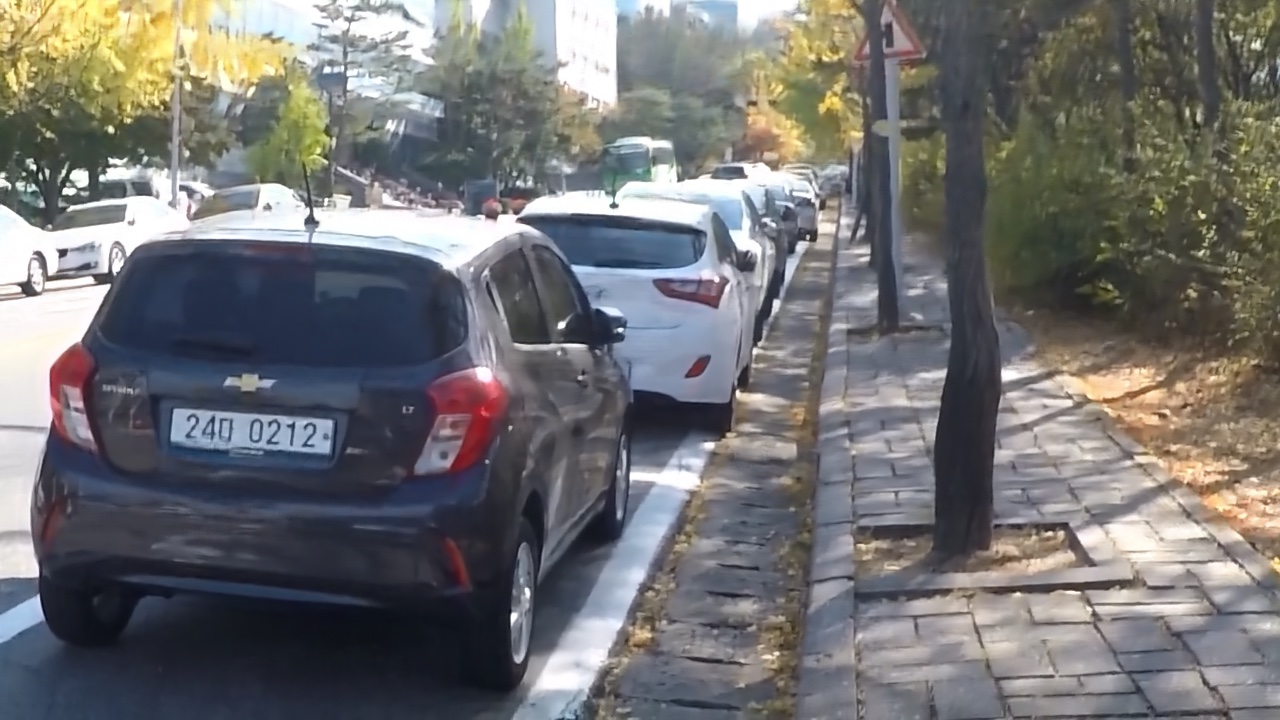} \hspace{\g} &
						\includegraphics[trim={158 230 935 378},clip,height=\h \textwidth, width=\w \textwidth]{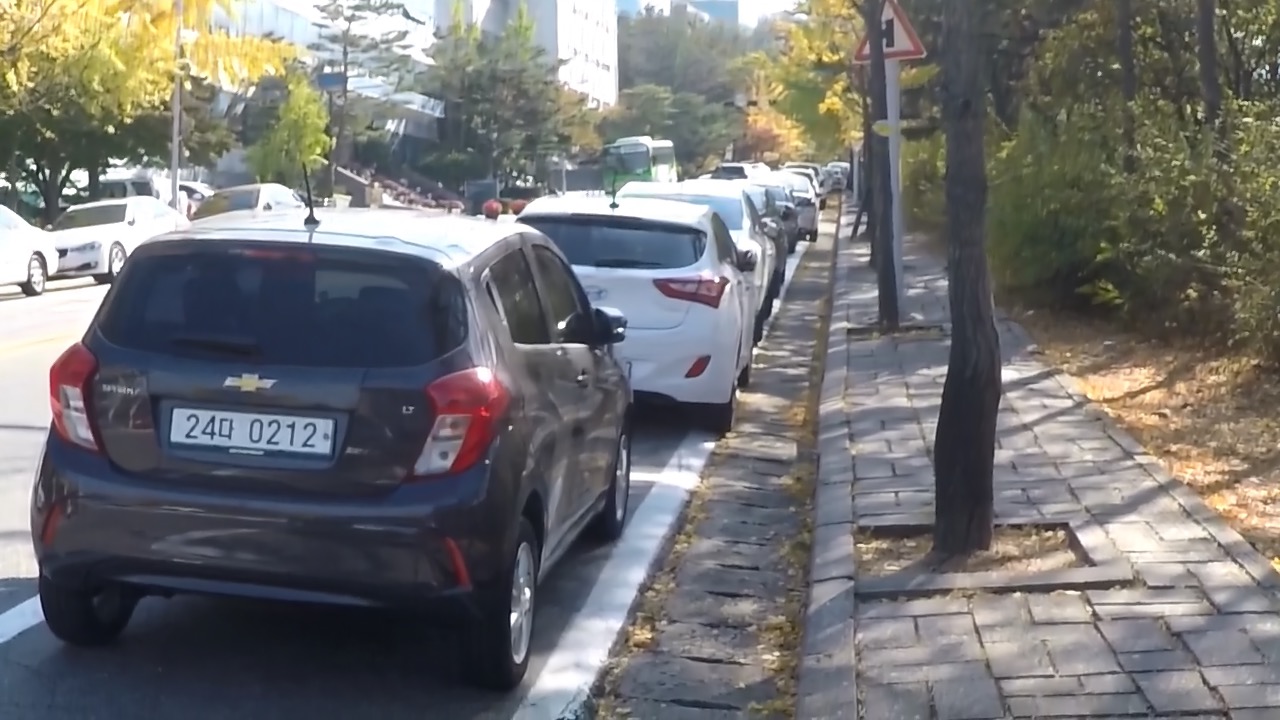}
						\hspace{\g} &		
						\includegraphics[trim={158 230 935 378},clip,height=\h \textwidth, width=\w \textwidth]{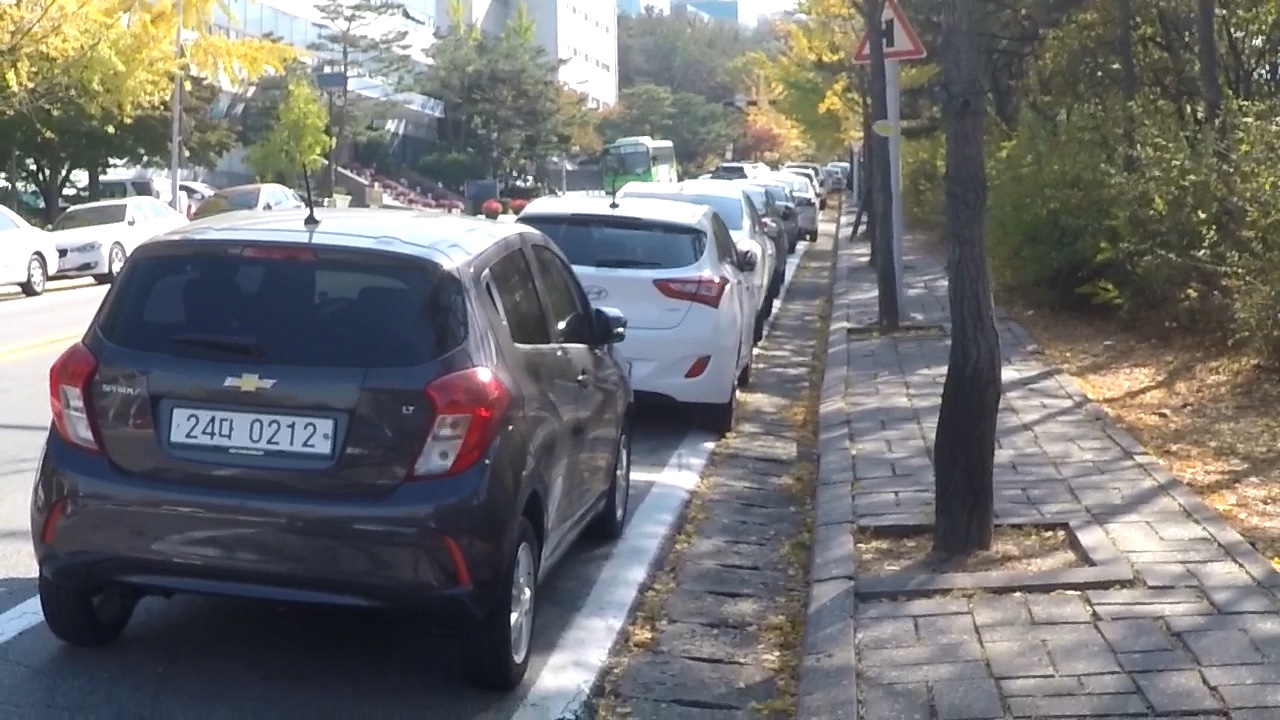} 
						\\ 
						
						SRN+~\cite{tao2018scale} & HINet+~\cite{chen2021hinet}  & \textbf{EFNet (Ours)}   & GT
						\\
					\end{tabular}
				\end{adjustbox}
				% \\ \large(a)
			\end{tabular}
			
			%	\\
			%	\vspace{2mm}
		\end{tabular}
	}
	\\ \vspace{2mm}

	\resizebox{1.00\linewidth}{!} {
		%   	\hspace{-2mm}
		\begin{tabular}{cc}
			
			\renewcommand{\name}{figures_sup/gopro/}
			\renewcommand{\h}{0.12}
			\renewcommand{\w}{0.2}
			\begin{tabular}{cc}
				%\tiny
				%\scriptsize
				%\footnotesize
				%\small
				% \hspace{-2.3mm}
				\begin{adjustbox}{valign=t}
					%\tiny
					\begin{tabular}{c}
		         	\includegraphics[trim={  0  0 416 0 },clip, width=0.354\textwidth]{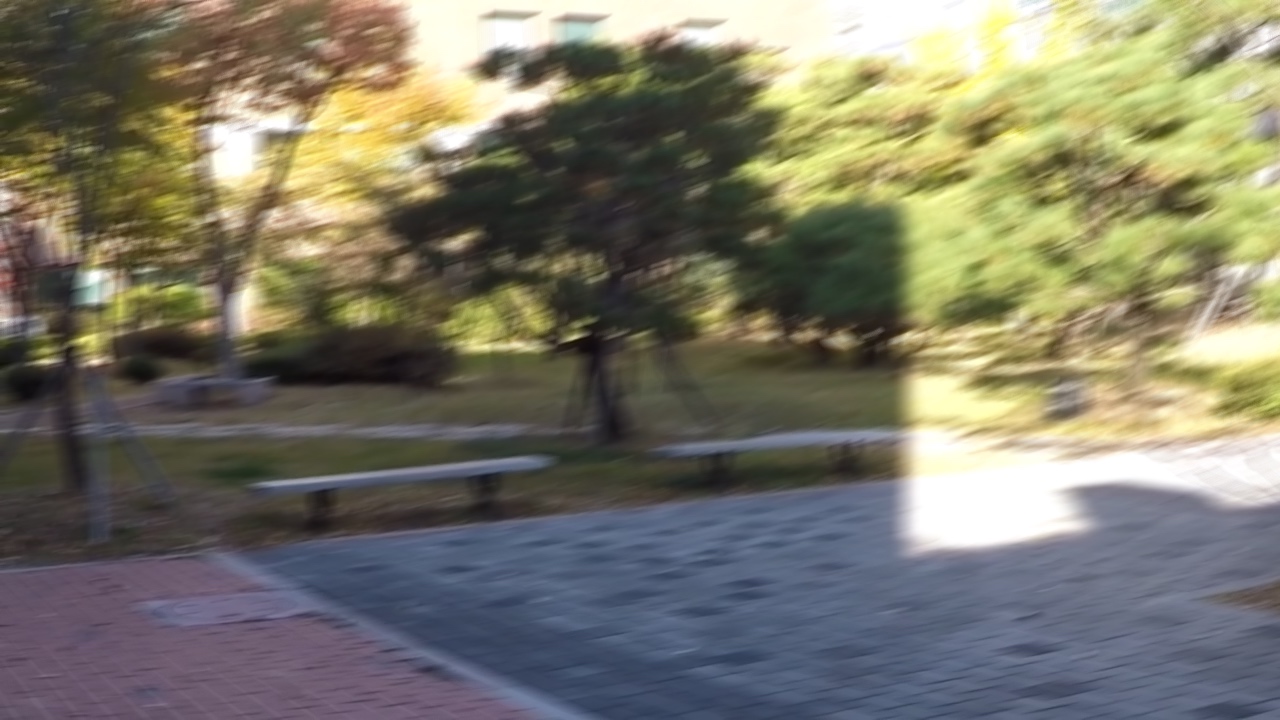}
						\\
						Blurry Image
					\end{tabular}
				\end{adjustbox}
				%		\hspace{-4.2mm}
				\begin{adjustbox}{valign=t}
					%\tiny
					\begin{tabular}{cccccc}
						\includegraphics[trim={48 15 928 520},clip,height=\h \textwidth, width=\w \textwidth]{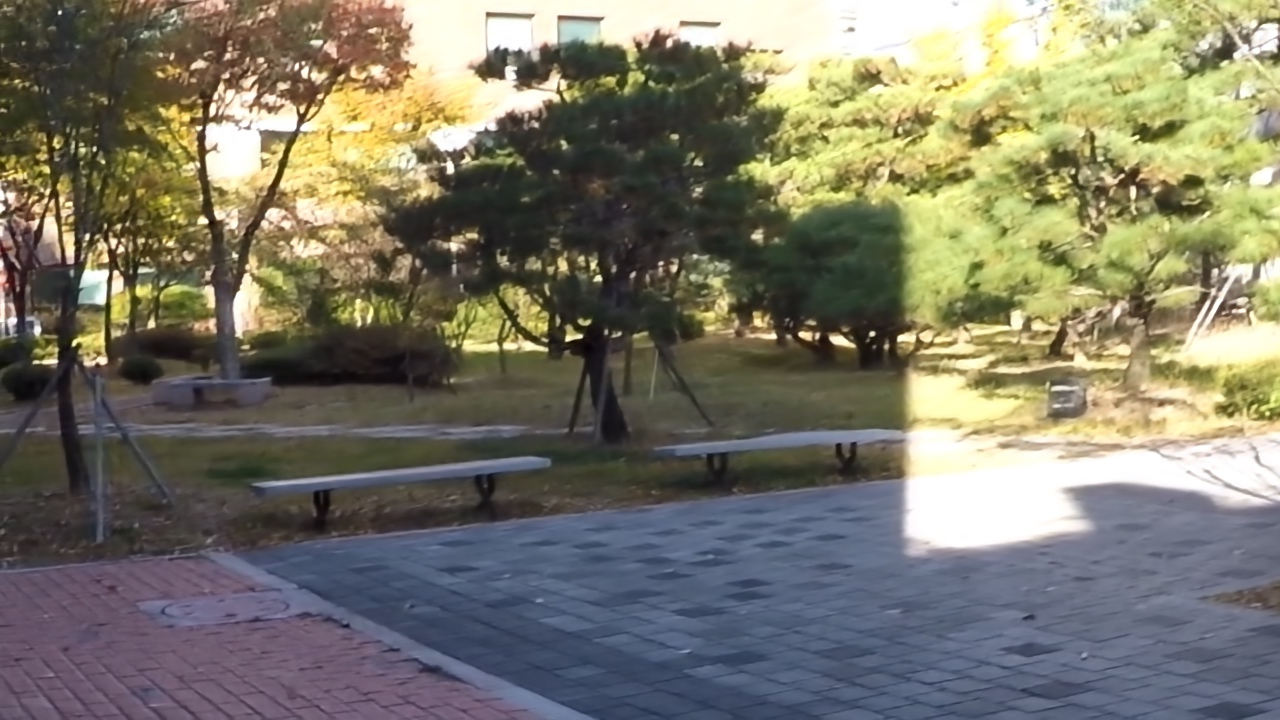} \hspace{\g} &
						\includegraphics[trim={48 15 928 520},clip,height=\h \textwidth, width=\w \textwidth]{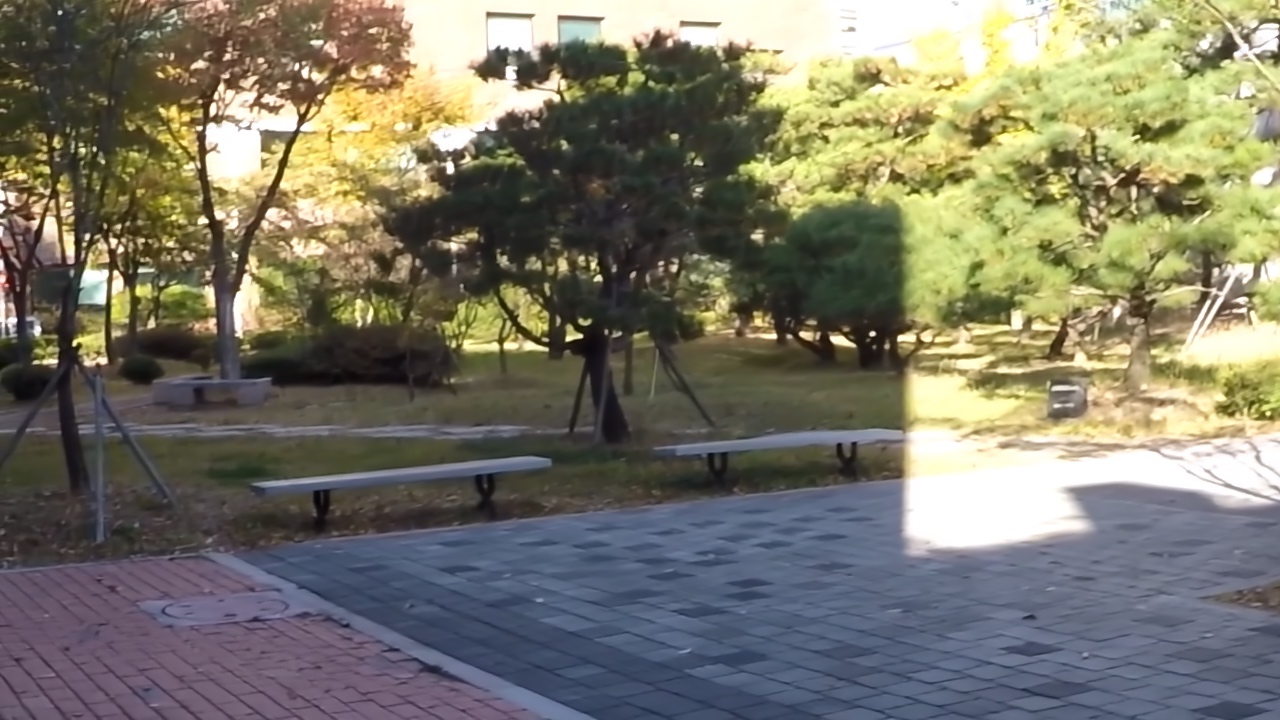} \hspace{\g} &
						\includegraphics[trim={48 15 928 520},clip,height=\h \textwidth, width=\w \textwidth]{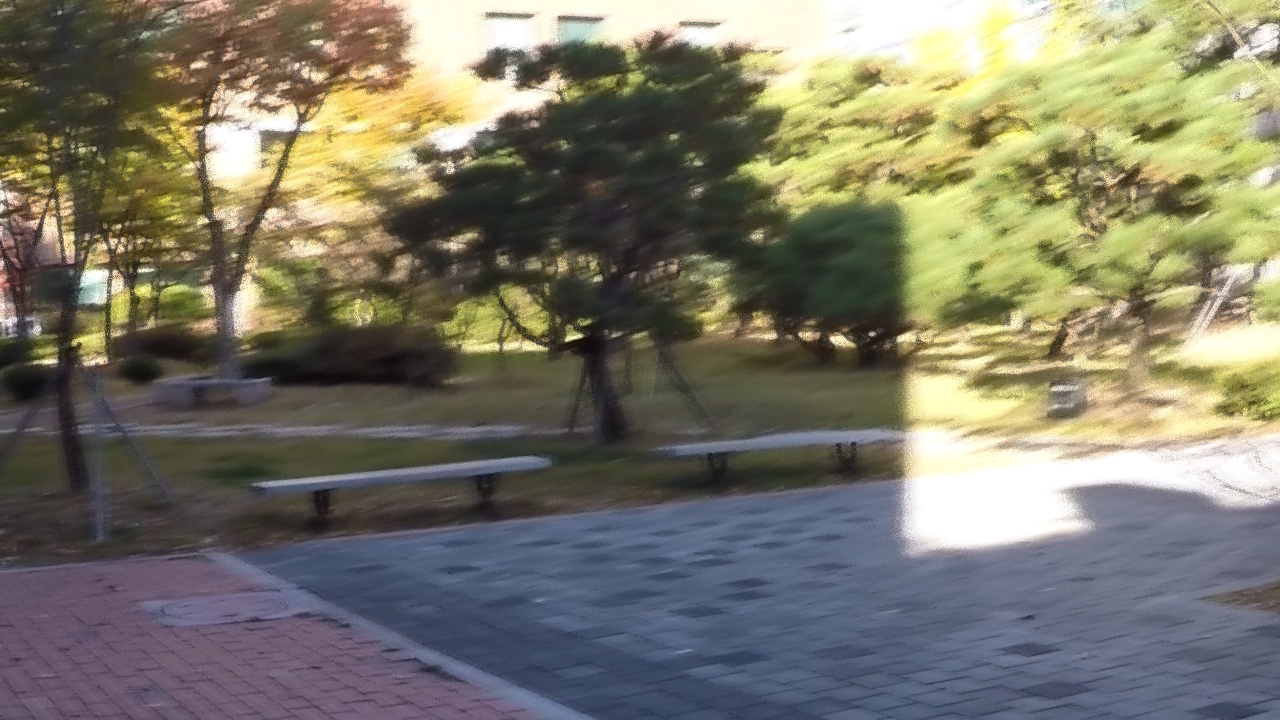} \hspace{\g} &
						\includegraphics[trim={48 15 928 520},clip,height=\h \textwidth, width=\w \textwidth]{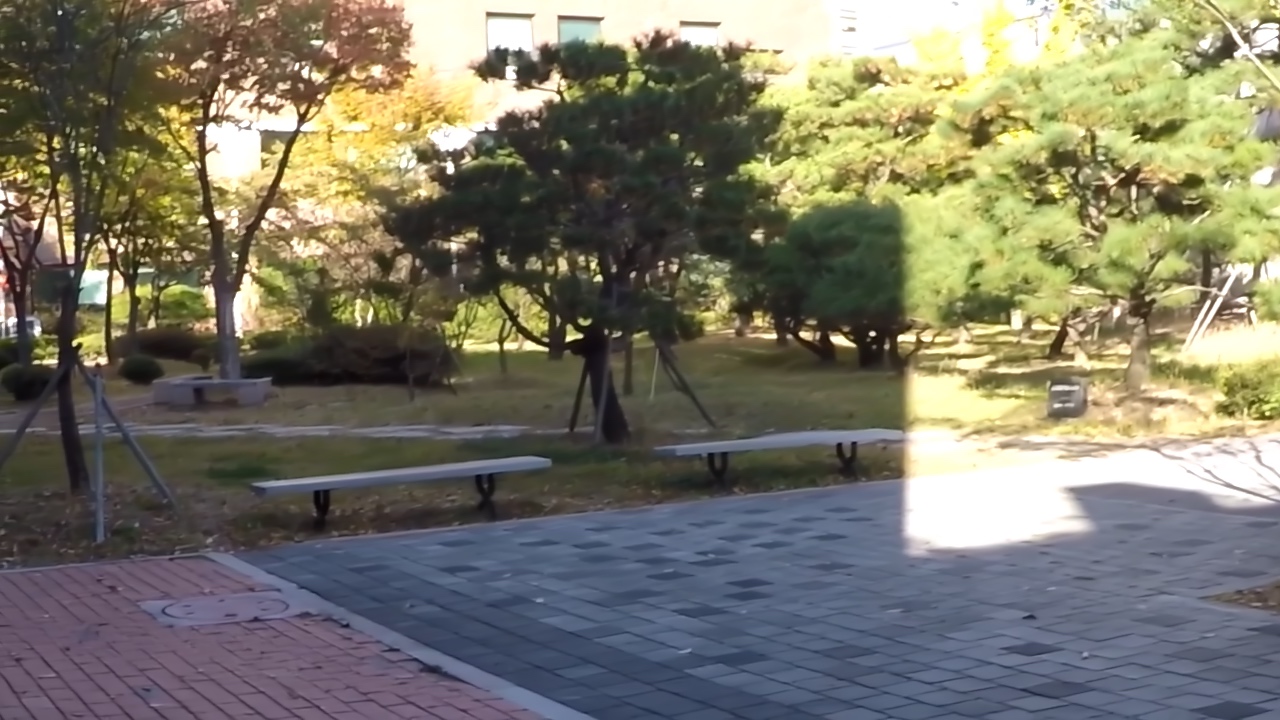} %MPRNet
						\\
						SRN~\cite{tao2018scale} &  HINet~\cite{chen2021hinet}  & BHA~\cite{pan2019bringing_high_framerate} &  MPRNet~\cite{zamir2021multi}
						\\
						\vspace{-2.45mm}
						\\
						
						\includegraphics[trim={48 15 928 520},clip,height=\h \textwidth, width=\w \textwidth]{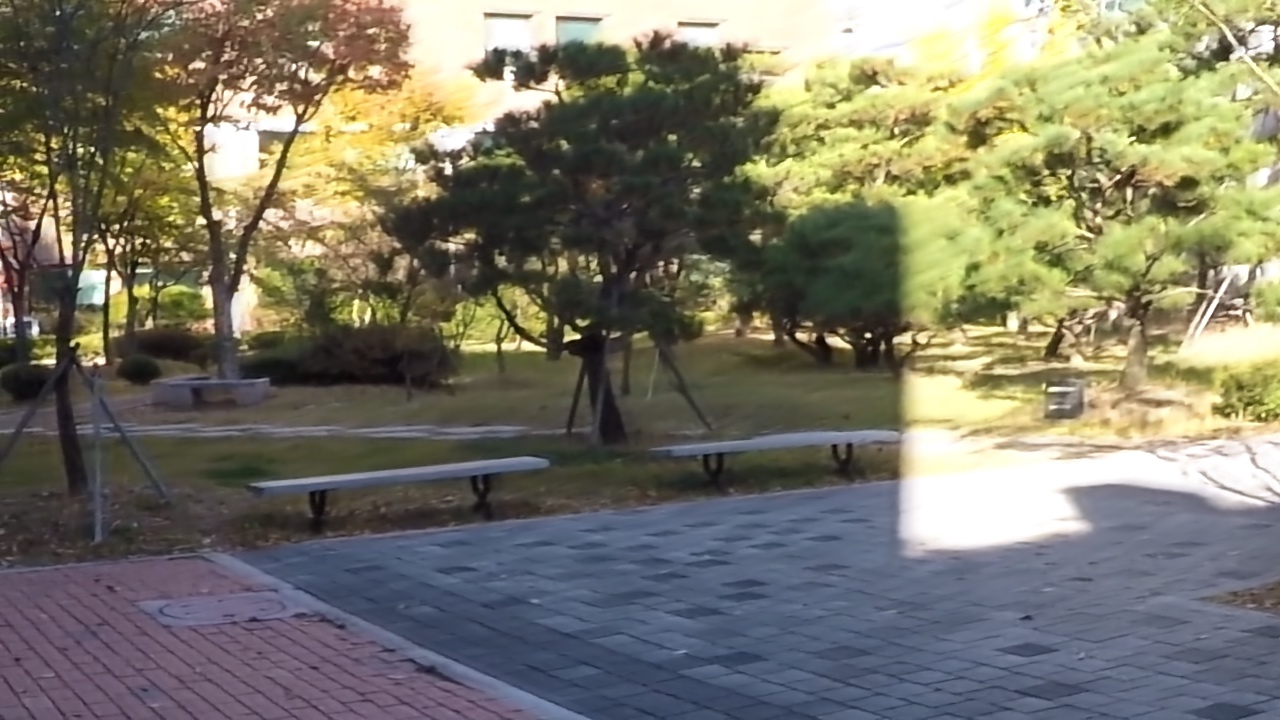} \hspace{\g} &
						\includegraphics[trim={48 15 928 520},clip,height=\h \textwidth, width=\w \textwidth]{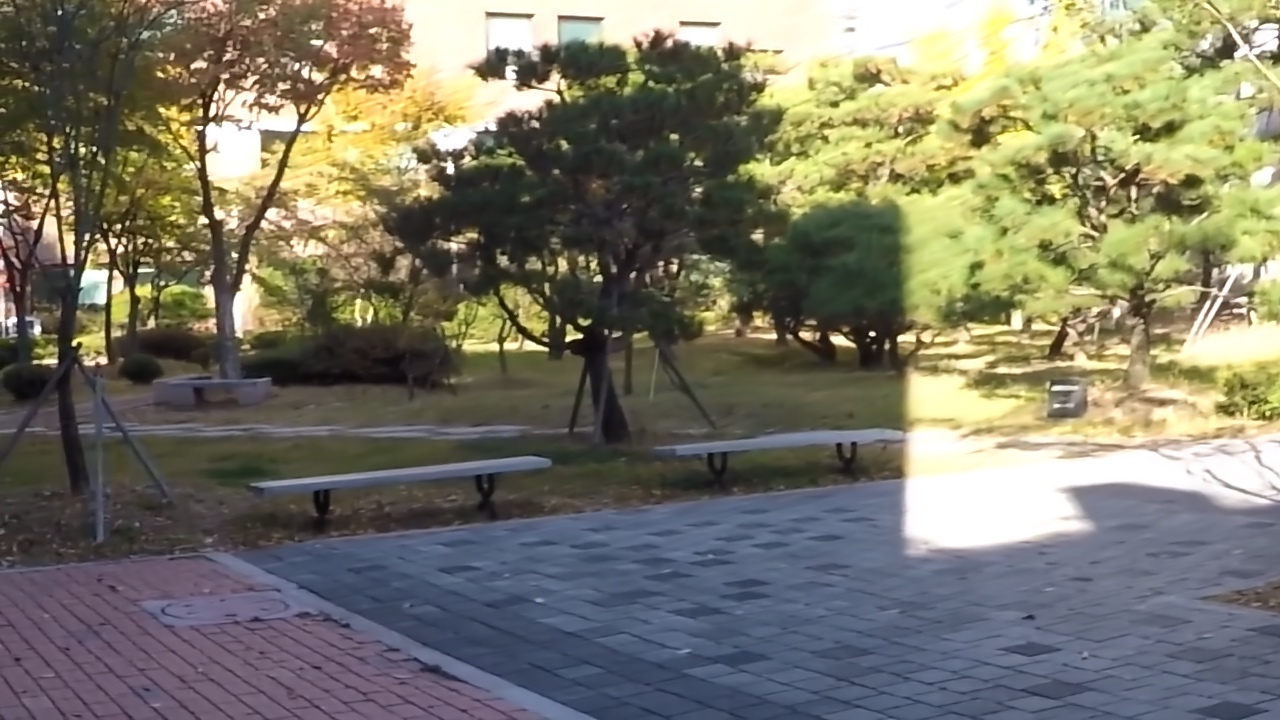} \hspace{\g} &
						\includegraphics[trim={48 15 928 520},clip,height=\h \textwidth, width=\w \textwidth]{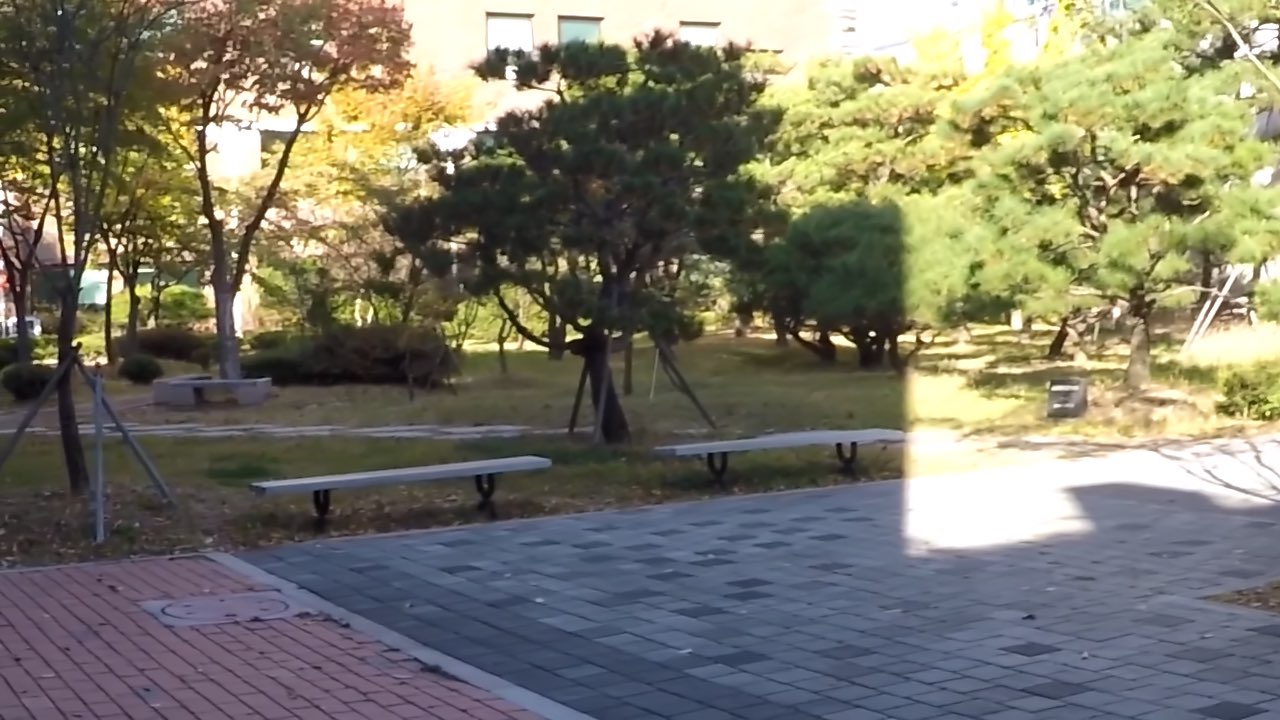}
						\hspace{\g} &		
						\includegraphics[trim={48 15 928 520},clip,height=\h \textwidth, width=\w \textwidth]{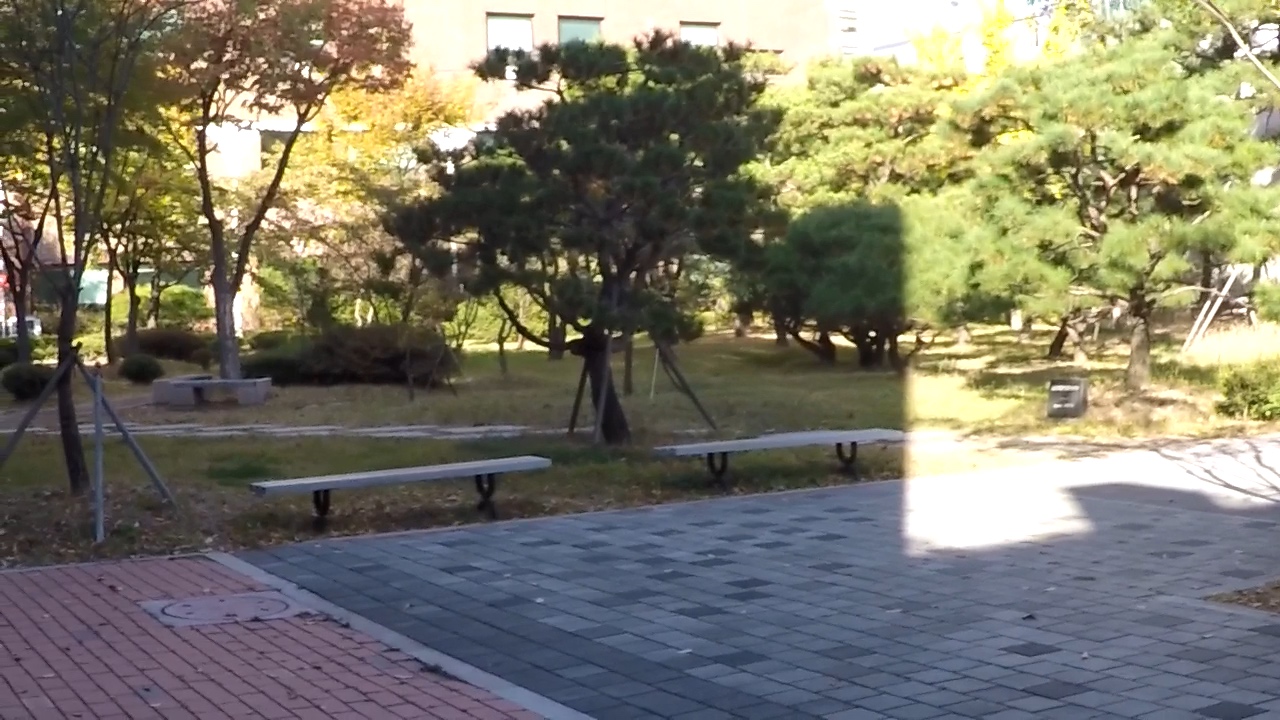} 
						\\ 
						
						SRN+~\cite{tao2018scale} & HINet+~\cite{chen2021hinet}  & \textbf{EFNet (Ours)}   & GT
						\\
					\end{tabular}
				\end{adjustbox}
			\end{tabular}
			
			%	\\
			%	\vspace{2mm}
		\end{tabular}
	}
	
	\vspace{-3mm}
	\caption{\textbf{Visual comparison on the GoPro dataset.} SRN+ and HINet+ are event-enhanced versions of SRN and HINet using our SCER. Compared to image-based and event-based state-of-the-art methods, our method restores fine texture and structural patterns better.}
% 	  \vspace{1mm}
	\label{fig:qualitative_gopro}
	% \vspace{-7mm}
\end{figure*}

\subsection{Comparisons with State-of-the-Art Methods}

We compare our method with state-of-the-art image-only and event-based deblurring methods on GoPro and REBlur. Since most learning-based methods using events do not have publicly available implementations, in the qualitative comparison part, apart from BHA~\cite{pan2019bringing_high_framerate}, we compare our method with SRN~\cite{tao2018scale} and HINet~\cite{chen2021hinet}, the latter being the current best model on the GoPro benchmark. To have a fair comparison, we also include event-enhanced versions of these two models by concatenating event voxel grids and images in the input.

\PAR{GoPro dataset.}
We report deblurring results in Table~\ref{table:gopro}. Compared to the best existing image-based method~\cite{chen2021hinet} and event-based method~\cite{haoyu2020learning}, our method achieves \textbf{2.75 dB} and \textbf{2.47dB} improvement in PSNR and 0.013 and 0.037 improvement in SSIM, resp., with a low parameter count of 8.47M. Despite utilizing an extra modality, other learning-based methods using events such as D$^{2}$Nets, LEMD, and ERDNet do not improve significantly upon image-only methods, indicating that they do not take full advantage of event features. Our model sets the new state of the art in image deblurring, showing that our principled two-stage architecture with multi-level attentive fusion leverages event information more effectively for this task. Note that by simply including our SCER to HINet~\cite{chen2021hinet}, the resulting enhanced version of it also surpasses the best previous event-based method~\cite{haoyu2020learning}. We show qualitative results on GoPro in Fig.~\ref{fig:qualitative_gopro}. Results from image-based methods are more blurry, losing sharp edge information. BHA~\cite{pan2019bringing_high_framerate} restores edges better but suffers from noise around them because of the factors described in Sec.~\ref{sec:method:theory}. Learning-based methods using events cannot fully exploit the motion information from events. By inputting the concatenation of our SCER with the image to SRN+ and HINet+, they both achieve large improvements. However, results from SRN+ include artifacts and noise due to the absence of a second stage in the network that would refine the result. HINet+ produces results with more artifacts, indicating that simply concatenating events and images in the input is not effective. Based on the physical model for event deblurring, EFNet achieves sharp and faithful results. Both dominant structures and fine details are restored well thanks to our attentive fusion at multiple levels.

\begin{table}[tb]
    \vspace{-3mm}
    % \captionsetup{font=small}
    \caption{\textbf{Comparison of motion deblurring methods on our REBlur dataset.} The notation is the same as in Table~\ref{table:gopro}.}
    % \vspace{-3mm}
	\centering
	\small
    \resizebox{0.95\textwidth}{!}{
    \setlength{\tabcolsep}{20pt}     % adjust the width of cell
    \renewcommand\arraystretch{1.0} % adjust hight of the row
        \begin{tabular}{ l  ||  c  |  c  |  c } 
		\bottomrule[0.15em]
        \rowcolor{tableHeadGray}
		Method & PSNR $\uparrow$ & SSIM $\uparrow$  & Params (M) $\downarrow$ \\ \hline \hline
		SRN~\cite{tao2018scale} & 35.10 (29.4\%) & 0.961 (35.9\%) & 10.25 \\
		HINet~\cite{chen2021hinet} & 35.58 (25.4\%) & 0.965 (28.6\%) & 88.67 \\
		BHA$^{\dagger}$~\cite{pan2019bringing_high_framerate} & 36.52 (16.8\%) & 0.964  (30.6\%) & 0.51 \\
		SRN+$^{\dagger}$~\cite{tao2018scale} & 36.87  (13.4\%) & 0.970 (16.7\%) & 10.43 \\
		HINet+$^{\dagger}$~\cite{chen2021hinet} & 37.68 (4.9\%) & 0.973 (7.4\%) & 88.85 \\ \hline
		\textbf{EFNet (Ours)$^{\dagger}$} & \textbf{38.12} & \textbf{0.975} & 8.47 \\  \hline
		
	\end{tabular}}
	% \vspace{-5mm}
	\label{tab:reblur_metrics}
\end{table}

\begin{figure*}[tb]
    % \captionsetup{font=small}
    \centering 
    \includegraphics[width=1.0\textwidth]{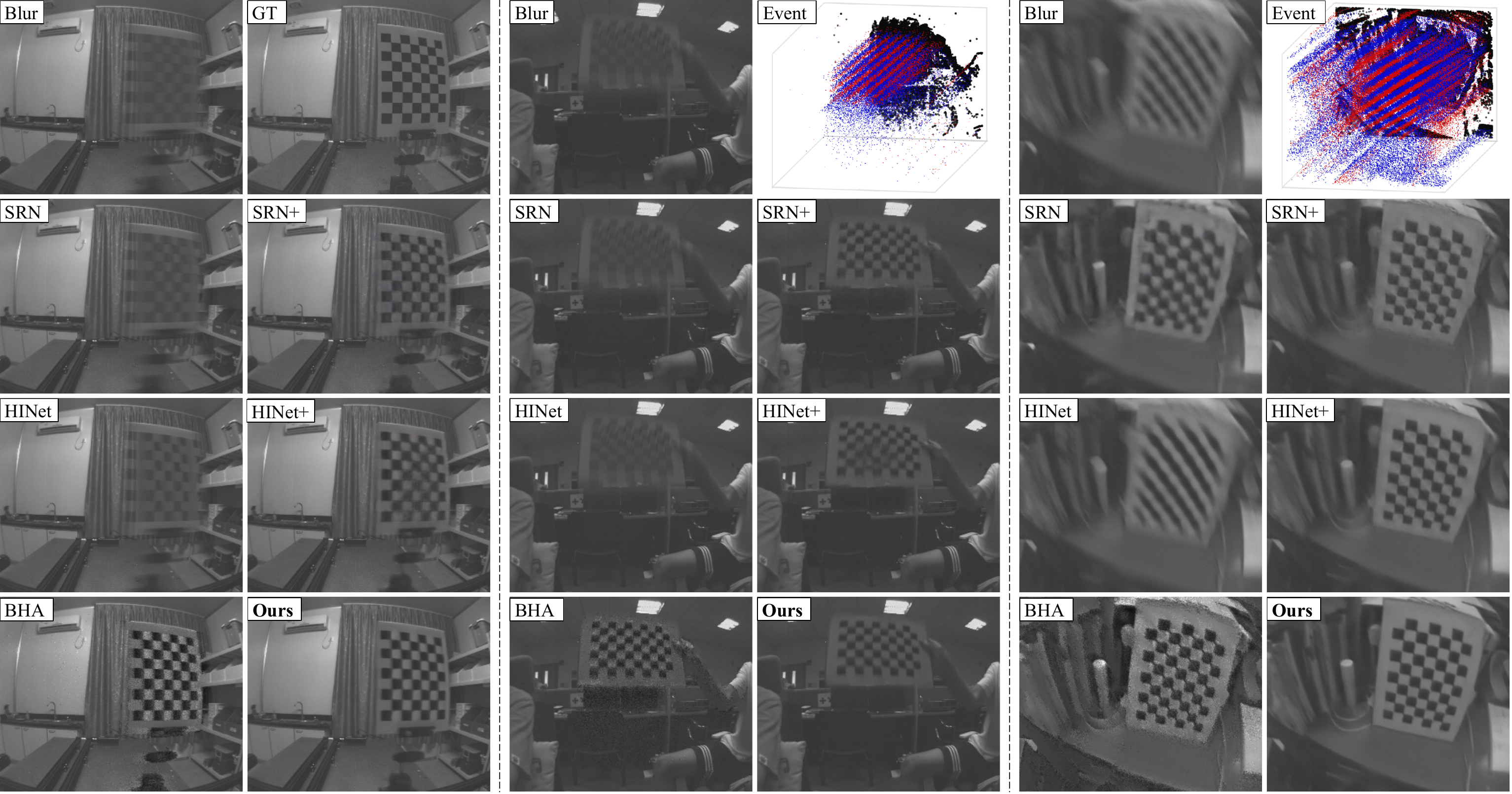}
    \vspace{-7mm}
    \caption{\textbf{Visual comparison on the REBlur dataset.} The first two columns are from the test set of the REBlur dataset, and the rest are from the additional set, for which ground truth is not available. Our method shows superior performance in cases with severe blur both due to object motion and due to camera motion. Best viewed on a screen and zoomed in.}
    \label{fig:reblur}
    \vspace{-5mm}
\end{figure*}

\PAR{REBlur dataset.}
We report quantitative results on REBlur in Table~\ref{tab:reblur_metrics}. Our model outperforms all other methods in this challenging real-world setting. Fig.~\ref{fig:reblur} depicts qualitative results from the test set and the additional set. Even the best image-based method, HINet, does not perform well on these severe cases of real-world motion blur. Event-based methods are more robust to such adverse conditions and less prone to overfitting on synthetic training data. Results from BHA are sharper, but accumulation noise still exists. Simply adding events with our SCER representation to the state-of-the-art image-based method~\cite{chen2021hinet} improves performance significantly because of the physical basis of SCER, but still leads to artifacts and ghost structures. Our EFNet restores both smooth texture and sharp edges, demonstrating the utility of our two-stage architecture and our cross-modal attention for fusion. Thanks to the selective feature connection via EMGC, EFNet restores blurry regions well while also maintaining the content of sharp regions. Results on more images are provided in the supplement.

\subsection{Ablation Study}

We conduct two ablation studies on GoPro to analyze the contribution of different components of our network (Table~\ref{tab:ablation_architecture}) and our event representation (Table~\ref{tab:ablation_representation}).
% \PAR{Fusion with EICA.}
First and foremost, our EICA fusion block fuses event and image features effectively, improving PSNR by 0.6 dB or more and SSIM by 0.4\% compared to simple strategies for fusion such as multiplication or addition (rows 6--8 and 10 of Table~\ref{tab:ablation_architecture}). Second, simply introducing middle fusion at multiple levels and using simple strategies for fusion yields an improvement of $\sim$1 dB in PSNR and 0.7\% in SSIM over early fusion (rows 5--8), evidencing the benefit of using multi-level fusion in our EFNet. Third, simply adding events as input to the network via early fusion of our SCER voxel grids with the images improves PSNR by 1.53 dB and SSIM by 0.6\% (rows 3--4), showcasing the informativeness of the event modality regarding motion, which leads to better deblurring results. Fourth, adding a second stage in our network to progressively restore the blurry image benefits deblurring significantly, both in the image-only case (rows 1 and 3) and in the case where our fully-fledged EFNet is used (rows 2 and 10). Fifth, connecting the two stages of EFNet with our EMGC improves the selective flow of information from the first stage to the second one, yielding an improvement of 0.15 dB (rows 9--10). Finally, all our contributions put together yield a substantial improvement of \textbf{6.4 dB} in PSNR and \textbf{3.6\%} in SSIM over the image-only one-stage baseline, setting the new state of the art in motion deblurring.

\begin{table}[tb]
    \vspace{-3mm}
    \caption{\textbf{Ablation study of various components of our method} on GoPro~\cite{nah2017deep}. ``Early'': fusion by concatenation of event voxel grid and image, ``Multi-level'': fusion with our proposed  architecture. SCER is used to represent events.}
    % \vspace{-1pt}
    \resizebox{1\textwidth}{!}{%
    \setlength{\tabcolsep}{8pt}     % adjust the width of blank space in the cell
    \renewcommand\arraystretch{1.1} % adjust hight of the row
    \begin{tabular}{cccccc||cc}
    \bottomrule[0.15em]
    \rowcolor{tableHeadGray}
     & Architecture & Events & Fusion type & EMGC &  Fusion module            & PSNR $\uparrow$ & SSIM $\uparrow$  \\ \hline \hline
    1 & 1-Stage        & \xmark                 & n/a & n/a             &    n/a     &    29.06  & 0.936\\
    2 & 1-Stage        & \cmark                 & Multi-level & n/a             &  EICA  & 34.90  & 0.968 \\ 
    3 & 2-Stage        & \xmark & n/a                & n/a        &  n/a      &    32.15   & 0.954 \\
    4 & 2-Stage        & \cmark & Early                & \xmark       &   n/a  &    33.68  & 0.960 \\
    5 & 2-Stage        & \cmark & Early                & \cmark       &   n/a  &    33.79  & 0.961  \\ % \hline
    6 & 2-Stage        & \cmark & Multi-level                & \cmark       &   Concat.  &    34.80  & 0.968  \\
    % 2-Stage        & \cmark & Multi-level                & \xmark       &   Affine  &    34.95  & 0.967  \\
    7 & 2-Stage        & \cmark & Multi-level                & \cmark       &   Multiply      &  34.86   & 0.968  \\
    8 & 2-Stage        & \cmark & Multi-level                & \cmark        &  Add       &    34.78   & 0.968  \\
    % 2-Stage        & \cmark & Multi-level                & \cmark       &   Affine$^{*}$  &  34.78   & 0.967  \\
    9 & 2-Stage        & \cmark & Multi-level                & \xmark       &   EICA  &   35.31	& 0.971 \\
    10 & 2-Stage        & \cmark & Multi-level                & \cmark      &    EICA     &    \textbf{35.46}   & \textbf{0.972}  \\ \hline
    \end{tabular}
    }
    \label{tab:ablation_architecture}
    \vspace{-3mm}
\end{table}

% simply concatenating the image with SCER at the input level. When using our multi-level fusion strategy, our AEIF module results in a 0.23dB and 0.31 dB improvement compared to simple multiplication and addition at multiple levels, resp.

% \PAR{Two-stage architecture with EMGC.}
% refines details of the deblurred image with events and addresses the noise introduced by the events. Thus, our architecture gives a significant performance boost of 1.14 dB in PSNR and 0.011 in SSIM compared to using a one-stage network.
% Besides, EMGC between the two stages improves the selective flow of information from the first stage to the second one, yielding an improvement of 0.14 dB.

\begin{wraptable}{l}{5.5cm}
    \vspace{-20pt}
    \caption{\textbf{Comparison between different event representations} on the GoPro~\cite{nah2017deep} dataset. ``Stack'': temporal accumulation of events in a single channel.}
    \vspace{-10pt}
    \resizebox{0.45\textwidth}{!}{
    \setlength{\tabcolsep}{8pt}     % adjust the width of cell
    \renewcommand\arraystretch{1.2} % adjust hight of the row
    \begin{tabular}{c||cc}
    \bottomrule[0.15em]
    \rowcolor{tableHeadGray}
    Event representation & PSNR $\uparrow$ & SSIM $\uparrow$  \\ \hline \hline
    None (image-only) & 32.15 & 0.954  \\
    Stack & 31.90 & 0.950 \\
    SBT~\cite{wang2019event} & 35.12 & 0.970 \\ \hline
    \textbf{SCER (Ours)} & \textbf{35.46} & \textbf{0.972} \\
    \hline
    \end{tabular}
    }
    \label{tab:ablation_representation}
    \vspace{-6mm}
\end{wraptable} 

\PAR{Event representation.} Introducing events can improve performance due to the high temporal resolution of the event stream, which provides a vital signal for deblurring. Table~\ref{tab:ablation_representation} shows a comparison between SCER and other event representations, including SBT~\cite{wang2019event}, which accumulates polarities in fixed time intervals. We use the same number of intervals (6) for SBT and SCER for a fair comparison. Based explicitly on physics, SCER utilizes event information for image deblurring (35.46dB) more effectively than SBT (35.12 dB). Note that simply accumulating all events across the exposure time (``Stack'') deteriorates the performance compared to not using events at all, which demonstrates that finding a suitable event representation for deblurring, such as SCER, is non-trivial.

% \begin{table}[]
%     \caption{\textbf{Comparison between different event representations} on the GoPro~\cite{nah2017deep} dataset. Image denotes feeding the image to the event branch. Stack denotes accumulating events in one channel. \LEI{transpose the table and use warp figure for one column table}}
%     \vspace{-4pt}
%     \resizebox{0.9\textwidth}{!}{
%     \setlength{\tabcolsep}{12pt}     % adjust the width of cell
%     \renewcommand\arraystretch{1.1} % adjust hight of the row
%     \begin{tabular}{c||cccc}
%     \bottomrule[0.15em]
%     \rowcolor{tableHeadGray}
%     Representation / Input of the event branch & Image & Stack  & SBT~\cite{wang2019event} Voxel grid~\cite{rebecq2019events} & \textbf{Ours} \\ \hline \hline
%     Simply concat. PSNR & xx    & xx & xx    &xx \\
%     SSIM & xxx & xx & xx & xx \\ \hline
    
%     Full model PSNR &  -  &   running    & - running  &   \textbf{35.09}    \\  
%     SSIM           & - &    running     & -  running &   \textbf{0.970}     \\\hline
%     \end{tabular}
%     }
%     \label{tab:ablation_representation}
%     \vspace{-6mm}
% \end{table}

\section{Conclusion}
\label{sec:conclusion}

In this work, we have looked into single image motion deblurring from the perspective of event-based fusion. Based on the common physical model which describes both blurry image formation and event generation, we have introduced EFNet, an end-to-end motion deblurring network with an attention-based event-image fusion module applied at multiple levels of the network. In addition, we have proposed a novel event voxel representation to best utilize events for deblurring. We have captured a new real-world dataset, REBlur, including several cases of severe motion blur, which provides a challenging evaluation setting. Our EFNet significantly surpasses the prior state of the art in image deblurring, both on the GoPro dataset and on our new REBlur dataset.

\PAR{Acknowledgements.} This work was partly supported by China Scholarship Council and Sunny Optical Technology (Group) Co., Ltd.

% \clearpage
\bibliographystyle{splncs04}
\bibliography{egbib}

\end{document}